\documentclass{article} 
\usepackage{iclr2027_conference,times}

\usepackage{amsmath,amsfonts,bm}

\def\eqref#1{equation~\ref{#1}}

\def\1{\bm{1}}

\DeclareMathAlphabet{\mathsfit}{\encodingdefault}{\sfdefault}{m}{sl}
\SetMathAlphabet{\mathsfit}{bold}{\encodingdefault}{\sfdefault}{bx}{n}

\usepackage{hyperref}
\hypersetup{hidelinks}
\usepackage{url}
\usepackage{graphicx}
\usepackage{booktabs}
\usepackage{array}
\usepackage{placeins}
\usepackage[rightcaption,raggedright]{sidecap} 
\usepackage{wrapfig}
\usepackage{tabularx}   
\usepackage{xcolor}     
\usepackage{colortbl}

\definecolor{SRRFullInk}{HTML}{243746}
\definecolor{SRRFullRule}{HTML}{ADBCC4}
\definecolor{SRRFullBand}{HTML}{E2EBEF}
\definecolor{SRRFullHeader}{HTML}{F0F4F5}
\definecolor{SRRFullBase}{HTML}{F6F7F8}
\definecolor{SRRFullTeal}{HTML}{147F79}
\definecolor{SRRFullSRRTint}{HTML}{E9F4F2}
\definecolor{SRRFullHARCInk}{HTML}{97651A}
\definecolor{SRRFullHARCTint}{HTML}{FCF5E9}

\newcommand{\SRRFullTableFont}{\fontsize{8}{9.3}\selectfont}

\definecolor{RTHeader}{HTML}{F0F2F5}
\definecolor{RTBlue}{HTML}{E7F0F7}
\definecolor{RTSand}{HTML}{F7F1E3}
\definecolor{RTInk}{HTML}{23384A}
\definecolor{RTRule}{HTML}{AAB5BF}

\title{Routing Drift Alone Does Not Diagnose \\Failure in Merged MoE LLMs}

\author{%
\textbf{Yuanyi Wang}$^{1}$\thanks{Equal contribution.}\, \footnotemark[2],
\textbf{Yanggan Gu}$^{1}$\footnotemark[1],
\textbf{Su Lu}$^{1}$,
\textbf{Guanghao Zhu}$^{1}$,
\textbf{Pengkai Wang}$^{1}$, \\
\textbf{Yifan Yang}$^{1}$,
\textbf{Congkai Xie}$^{2}$,
\textbf{Zhaoyi Yan}$^{2}$,
\textbf{Jianmin Wu}$^{2}$,
\textbf{Hongxia Yang}$^{1,2}$\thanks{yuan-yi.wang@connect.polyu.hk, hongxia.yang@polyu.edu.hk} \\
$^{1}$The Hong Kong Polytechnic University, PolyU \quad
$^{2}$InfiX.ai
}

\iclrfinalcopy 
\begin{document}

\maketitle
\lhead{Routing Drift Alone Does Not Diagnose Failure in Merged MoE LLMs}

\begin{abstract}
Model merging efficiently combines specialized large language models (LLMs) without joint
retraining, but can substantially alter expert routing in Mixture-of-Experts
(MoE) models.
Such \emph{routing drift} is often interpreted as routing failure, raising a fundamental question that remains unclear:
\emph{does routing drift after MoE merging actually indicate routing failure, and what evidence
should justify repair?}
We investigate these questions across DeepSeekMoE, OLMoE, and Qwen3-MoE proposing
a routing analysis toolkit for controlled counterfactual interventions and
token-level analysis.
By crossing source and merged router inputs and parameters, we attribute most expert 
reassignments to input shifts rather than parameter changes at the same layer.
However, source-relative routing differences poorly predict next-token likelihood
gains from source-route restoration, and different expert selections can
produce directionally similar mixture outputs.
We therefore operationalize routing failure as 
\textit{task loss recoverable under a
specified routing intervention, with non-routing parameters fixed.}
These tests detect recoverable loss under deliberate router corruption,
whereas source-route restoration does not establish reliable task benefits
in the evaluated merged models.
Motivated by these, we propose \emph{Selective Router Repair (SRR)} as a case study,
and find that source-specialist token-likelihood
advantages do not reliably identify beneficial local corrections.
Together, these findings show that \textbf{routing drift alone is insufficient
evidence of routing failure}: source-informed corrections must be judged by
their task-level intervention effects.
The analysis toolkit and SRR code are released.
\footnote{Our toolkit and code are available at \url{https://github.com/wyy-code/SRR}.}
\end{abstract}

\section{Introduction}
Model merging combines complementary capabilities from independently specialized large language models (LLMs) without joint retraining
\citep{zhou2025democratizing,yang2026model,li2026model}.
Most existing methods target dense LLMs, where all tokens use the same feed-forward modules
\citep{wang2025model,zhou2026model}.
Mixture-of-Experts (MoE) models instead route each token to only a small subset of experts
\citep{jacobs1991adaptive,shazeer2017outrageously,lepikhin2020gshard}.
Merging MoEs can therefore change not only model parameters but also token-to-expert routing.
These changes in expert assignment and routing probabilities are referred to as \emph{routing drift}.

Routing drift is naturally concerning: if a merged model sends a token to different experts than its corresponding specialized source model, the router may appear to have failed, motivating recent routing-realignment methods \citep{huang2026model}.
However, a changed route is not necessarily a \emph{routing failure}.
We use \emph{routing failure} to denote degradation in task-relevant behavior attributable to routing, rather than a structural change in routing itself.
Whether routing drift reliably diagnoses such failure still remains unclear.
This raises our central question: \emph{does routing drift after MoE merging actually indicate routing failure, and what evidence should justify repair?}

\begin{figure*}[t]
  \centering
  \includegraphics[width=0.98\textwidth]{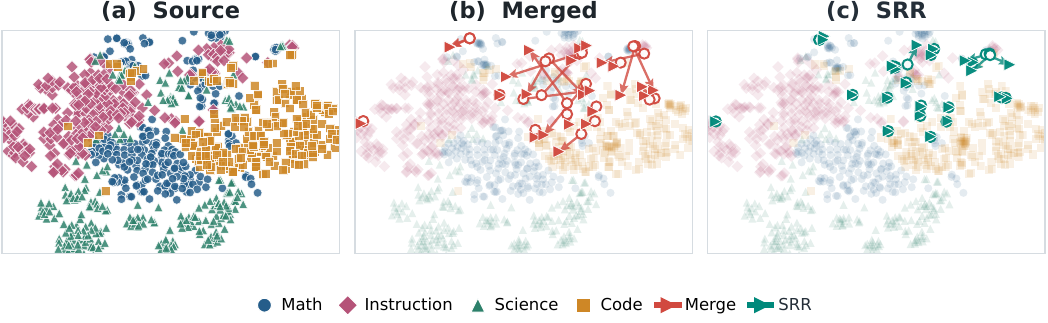}
  \vspace{-0.9em}
  \caption{\textbf{Merge-induced representation drift and SRR updates.}
  Illustrative joint t-SNE of aligned router-input representations in OLMoE.
  Arrows track the same domain-balanced tokens from Source to Merged in (b) and their corresponding shifts after SRR in (c).}
  \label{fig:representation-drift}
  \vspace{-1em}
\end{figure*}

Answering this question is nontrivial because routing drift alone reveals neither why routing changed nor whether it caused harm.
A token's route depends jointly on its router-input representation, the hidden state presented to the router, and the router parameters, so the same routing change may arise from representation shifts, router changes, or both.
Moreover, expert assignment is only an intermediate structural decision: the MoE layer output also depends on the weighted outputs of the selected experts.
Figure~\ref{fig:representation-drift}(a,b) illustrates post-merge representation
shifts for aligned tokens.
In short, routing drift therefore describes \emph{what changed}; routing failure asks \emph{what harm routing caused}.
Thus, establishing failure requires isolating the causal effect of routing.

To address these questions, we develop a routing analysis toolkit for controlled
counterfactual interventions and paired token- and task-level evaluation.
Across DeepSeekMoE, OLMoE, and Qwen3-MoE under different merging methods, 
we cross source and merged router inputs and parameters on aligned
token sequences.
In $77.7$--$96.9\%$ of changed-route events, input-only replacement changes
the source expert set, whereas parameter-only replacement does not.
This local attribution does not exclude upstream routing effects.
Yet full-distribution JS divergence poorly predicts next-token
likelihood gains from source-route restoration (under $0.52$ AUROC).
At fixed merged states and expert parameters, source and native merged routers
also produce similar mixtures, exceeding matched random-route
controls in cosine similarity.
Task consequences therefore require a separate test.

To provide that test, we operationalize routing failure as task loss recoverable
under an alternative routing policy, with non-routing parameters fixed.
We compare policies on identical items using the merged model's experts,
while allowing downstream representations to respond.
As a positive control, replaying clean routes after permuting OLMoE
router logits recovers an accuracy loss.
By contrast, source-route restoration in evaluated merged models does not
establish reliable task benefits.
An inconclusive result establishes neither harmlessness nor optimal routing.
This grounds the diagnosis in task consequences relative to a specified
intervention, not source agreement.

This criterion also lets us examine whether source predictions justify selective
corrections.
A source specialist's token-likelihood advantage compares models,
rather than isolating the value of a routing update.
We propose \emph{Selective Router Repair (SRR)} as a case study: it fits
likelihood-weighted, source-derived expert-pair corrections on merged router
inputs, updating selected router parameters.
Figure~\ref{fig:representation-drift}(c) illustrates a candidate's
representation changes, not task recovery.
On independent diagnostic prompts, we compare source-informed and fitted
directions with native and opposite-direction controls within the merged model.
These tests do not establish that positive source-likelihood advantage selects
beneficial local corrections; paired evaluation on fixed checkpoints does not
establish an average task gain.
Together, these findings show that \textbf{routing drift alone is insufficient
evidence of routing failure}: candidate construction and task
recovery are distinct claims.

In summary, our contributions are summarized as follows:
\\
\textbf{(1) Routing drift analysis:}
Across different MoE models, local interventions attribute most reassignments
to input shifts, while structural differences poorly predict source-route
intervention gains.
\\
\textbf{(2) Task-grounded diagnosis:}
We designed analysis toolkit to formalize intervention-relative recoverable task loss 
and implement paired
routing tests with fixed non-routing parameters.
\\
\textbf{(3) Repair supervision:}
Our proposed SRR finds no reliable evidence that source-likelihood advantages
identify beneficial corrections or that fitted updates improve
performance on fixed checkpoints.

\section{Related Work}
\textbf{Model Merging} combines specialized models without joint retraining or ensemble inference.
Prior work spans parameter averaging and weighted combination
\citep{wortsman2022model,matena2022merging},
task-vector composition
\citep{ilharco2022editing},
and interference reduction through sign resolution
\citep{yadav2023ties},
sparsification
\citep{yu2024language},
adaptive coefficients
\citep{yang2024adamerging,stoica2024zipit,wang2026discovering},
feature alignment
\citep{lu2024twin,stoica2025model,wang2026pmq},
and low-rank structure
\citep{gargiulo2025task,cheng2025whoever}.
Recent studies scale merging to LLMs
\citep{akiba2025evolutionary,wang2026geometry},
automate merge search
\citep{wang2026mergepipe},
study scaling laws
\citep{wang2025model},
and extend merging to agents and embodied models
\citep{yuan2026behavior,fu2026mergevla,li2025deep}.
These studies primarily ask \emph{how} dense model parameters should be combined. 
Our focus is complementary:
we investigate what routing drift mean within an already
merged MoE and what evidence should justify repair.

\textbf{Routing Drift and Repair.}
Sparse MoEs use learned routers to select token-specific experts
\citep{fedus2022switch,dua2022tricks,zoph2022st,zhou2022mixture,puigcerver2024sparse},
with modern MoE LLMs increasingly relying on large-scale expert specialization
\citep{jiang2024mixtral,xu2026deepseek,liu2024deepseek,yang2025qwen3}.
Routing has also guided expert merging and compression
\citep{li2024merge}.
Most closely related, HARC treats source-to-merged routing mismatch as routing breakdown and realigns the merged router
\citep{huang2026model}.
Other post-merge methods instead align representations or features, including Surgery 
\citep{yang2024representation}, ProbSurgery
\citep{wei2025representation}, FeatCal
\citep{gu2026featcal}, and Expert Merging
\citep{zhang2026expert}.
Our work instead asks whether routing change constitutes failure.
We distinguish alignment as a repair objective from disagreement as evidence
of routing failure.
Our toolkit separates local input and router-parameter effects, then measures
recoverable task loss under specified routing interventions with non-routing
parameters fixed.
Proposing SRR as a case study, we further test whether source-likelihood advantages
support useful local corrections, rather than treating them as failure labels.

\section{Routing Drift Is Not Routing Failure}
\label{sec:routing-drift}

In this section, we develop controlled intervention analysis that disentangles representation-
and router-induced changes.
We examine the origin of routing drift (Sec.~\ref{sec:route-origin}),
whether it predicts intervention gains (Sec.~\ref{sec:route-diagnosis}),
and how different routes preserve similar mixture outputs
(Sec.~\ref{sec:mixture-redundancy}).
We then formulate and validate task-grounded tests of routing failure
(Sec.~\ref{sec:task-grounded-failure}).

\subsection{Most Post-Merge Route Changes Are Representation-Induced}
\label{sec:route-origin}

\begin{figure*}[h]
  \centering
  \includegraphics[width=0.9\textwidth]{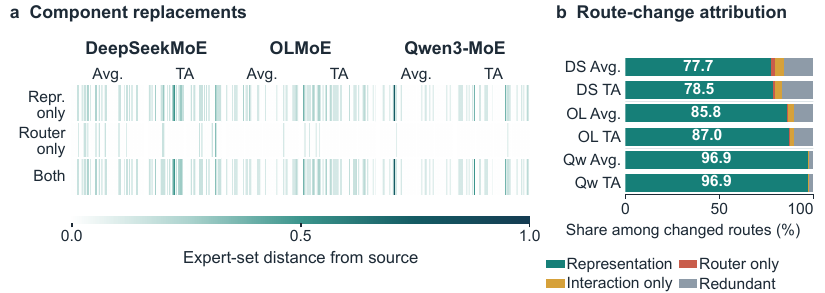}
  \vspace{-0.5em}
  \caption{\textbf{Most post-merge route changes are representation-induced.}
  (a) Expert-set distances for 64 aligned events per setting under component replacements.
  (b) Full-sample attribution among changed routes.
  Protocols and intervals appear in Appendices~\ref{app:cross-architecture-attribution} and~\ref{app:qwen-route-diagnosis}.}
  \label{fig:route-origin}
  \vspace{-0.6em}
\end{figure*}

Figure~\ref{fig:representation-drift}(a,b) illustrates shifts in router-input
representations; aligned input and output views across three architectures
appear in Appendix~\ref{app:illustrative-representation-trajectories}.
We cross source and merged representations with source and merged router parameters on identical token sequences.
Across DeepSeekMoE~\citep{dai2024deepseekmoe}, OLMoE~\citep{muennighoff2025olmoe}, and Qwen3-MoE~\citep{yang2025qwen3}, each under Average and Task Arithmetic merging, we analyze 256 domain-balanced prompts per setting, all continuation tokens, and all sparse layers.
Figure~\ref{fig:route-origin}(a) displays $1-|S_{\mathrm{src}}\cap S_{\mathrm{cond}}|/k$, where the sets contain the $k$ selected experts under source and replacement conditions.
Across these six settings, $26.5$--$54.3\%$ of token--layer expert sets change.
Among changed routes, $77.7$--$96.9\%$ change under representation-only replacement but not router-only replacement; we call these \emph{representation-induced}.
The converse, \emph{router-parameter-induced} changes, accounts for at most $1.8\%$ (Figure~\ref{fig:route-origin}(b)).
These categories do not require reproducing the merged expert set; full definitions and domain and details appear in Appendices~\ref{app:representation-routing-design}, \ref{app:cross-architecture-attribution}, and~\ref{app:qwen-route-diagnosis}.
Altered router inputs thus explain most observed expert reassignments, without establishing whether they are harmful.

\subsection{Routing Drift Poorly Predicts Source-Route Gains}
\label{sec:route-diagnosis}

\begin{figure*}[t]
  \centering
  \includegraphics[width=0.92\textwidth]{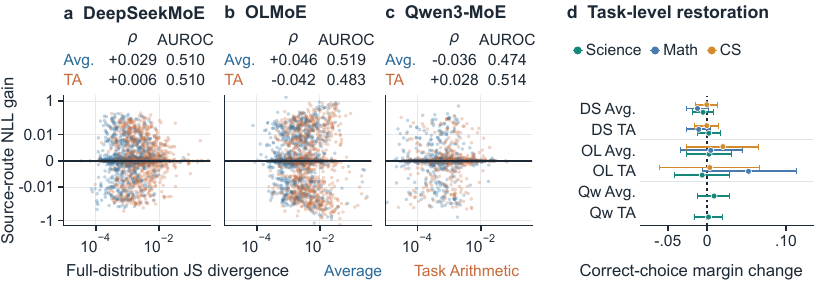}
  \vspace{-0.5em}
  \caption{\textbf{Routing drift alone is insufficient evidence of routing failure.}
  (a--c) JS divergence versus source-route NLL gain (log-$x$, symmetric-log $y$).
  (d) Task-margin changes with 95\% bootstrap intervals.
  Full tests appear in Appendices~\ref{app:routing-metric-family}--\ref{app:qwen-route-diagnosis}.}
  \vspace{-0.6em}
  \label{fig:route-diagnosis}
\end{figure*}

We test whether structural differences predict intervention gains: reductions in next-token negative log-likelihood (NLL) on fixed continuations.
On changed-route events in the final five sparse layers, positive and negative source-route gains occur at overlapping full-distribution Jensen--Shannon (JS) divergences across all three architectures (Figure~\ref{fig:route-diagnosis}(a--c)).
JS-based prediction remains near chance (AUROC $0.47$--$0.52$; chance: $0.5$).
Other distances, including Top-$k$ set distance, and intervention targets are reported in Appendices~\ref{app:routing-metric-family} and~\ref{app:qwen-route-diagnosis}.

We then restore source-derived expert selections and native routing weights while keeping model parameters fixed.
Token-local controls can increase NLL.
For task evaluation, we replace routes throughout question--choice sequences and score answer tokens only.
Changes in correct-choice margin---the correct option's score advantage over the strongest alternative---have 95\% confidence intervals spanning zero (Figure~\ref{fig:route-diagnosis}(d)), providing no reliable evidence of task improvement.
Accuracy results and protocols appear in Appendices~\ref{app:task-route-controls} and~\ref{app:qwen-route-diagnosis}.

This does not imply optimal merged routing.
Local tests on DeepSeekMoE and OLMoE show that the merged route is rarely best within fixed candidate sets, yet the source route wins only roughly half the comparisons (Appendix~\ref{app:local-reference-quality}).
Thus, \textbf{routing drift alone is insufficient evidence of routing failure},
the tested structural metrics poorly predict the benefit of
source-route restoration in these settings.
These comparisons neither establish optimal merged routing nor exclude
task loss recoverable under other routing policies.

\subsection{Different Routes Can Preserve Similar Mixture Outputs}
\label{sec:mixture-redundancy}

We compare source-route and native merged-route mixtures at the same merged hidden state $h$, with expert parameters fixed.
At layer $\ell$,
\begin{equation}
    o^{(\ell)}(h;S,g)
    = \sum_{e\in S} g_e E_e^{(\ell)}(h),
    \label{eq:routed-mixture}
\end{equation}
where $S$ is the selected expert set, $g_e$ the native weight of expert $e$, and $E_e^{(\ell)}(h)$ its output.

For each changed-route event, we compute the maximum output cosine between entering and leaving experts.
Across the six settings, its mean is $0.041$--$0.096$, whereas source-route and native merged-route mixtures have mean cosine $0.896$--$0.976$ (Figure~\ref{fig:mixture-redundancy}(a)).
Mixture cosine exceeds the expert-pair maximum in all 727 sampled events (Figure~\ref{fig:mixture-redundancy}(c)).

To control for overlap and weighting, we compare each source route with 32 random alternatives matched for expert count, overlap with the merged route, and routing-weight values.
Observed routes exceed these controls in mean cosine by $0.112$--$0.147$, with positive paired 95\% prompt-bootstrap intervals in every setting (Figure~\ref{fig:mixture-redundancy}(b); Appendix~\ref{app:changed-route-redundancy}).
This supports \emph{mixture-level functional redundancy}: different expert selections can preserve aligned mixture outputs, without guaranteeing unchanged magnitudes or downstream behavior.

\begingroup
\renewcommand{\sidecaptionsep}{0.03\textwidth}
\sidecaptionvpos{figure}{c}
\begin{SCfigure}[0.3][t]
  \includegraphics[width=0.7\textwidth]{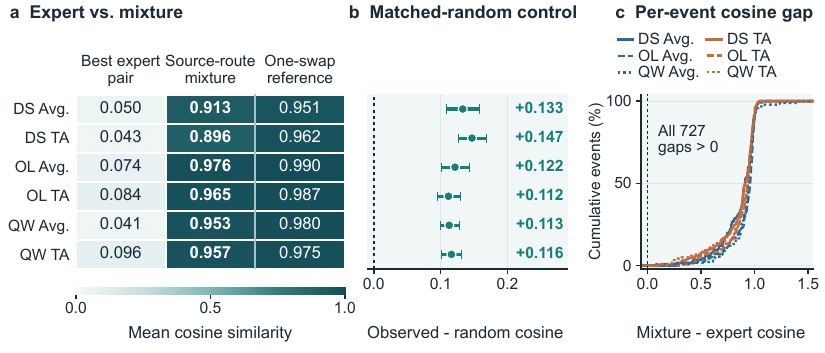}
  \caption{\textbf{Different routes, similar mixture outputs.}
(a) Mean cosines.
(b) Paired observed-minus-random cosine (95\% CIs).
(c) Cumulative event-wise cosine gaps.}
\vspace{-1em}
\label{fig:mixture-redundancy}
\end{SCfigure}
\endgroup

\section{Task-Grounded Diagnosis of Routing Failure}
\label{sec:task-grounded-failure}

\subsection{Intervention-Relative Recoverable Loss}
\label{sec:recoverable-routing-loss}

Section~\ref{sec:routing-drift} separates routing changes from their local
consequences. To test task harm, we compare a baseline routing policy $\pi$
with a specified alternative $\pi'$ within the same model.
Each policy determines expert selections and weights within a fixed
intervention scope. With non-routing parameters $\phi$ held fixed, define
\begin{equation}
H_{\mathcal D}(\pi;\pi')
=
\mathbb E_{(x,y)\sim\mathcal D}
\left[
\ell_{\mathrm{task}}(M_{\phi,\pi};x,y)
-
\ell_{\mathrm{task}}(M_{\phi,\pi'};x,y)
\right].
\label{eq:task-grounded-routing-harm}
\end{equation}
The alternative policy is part of the estimand: schedule replay, an online
router swap, and a calibrated router are distinct interventions because
they can induce different downstream states.
Both policies are evaluated on identical task items, allowing downstream
representations to respond to the intervention. For zero--one loss,
$H_{\mathcal D}$ equals the accuracy gain; positive values indicate
recoverable loss relative to the chosen task, alternative, and scope.
A confidence interval wholly above zero supports this recovery; an interval
containing zero establishes neither harmlessness nor optimality.
Recovery through routing does not by itself identify router-parameter changes as the
original cause of degradation.
Native-routing checks, scoring, and uncertainty procedures appear in
Appendix~\ref{app:task-grounded-failure}.

\subsection{Controlled Recovery and Natural Routing Alternatives}
\label{sec:diagnostic-controls}

\begin{wrapfigure}{r}{0.50\columnwidth}
\vspace{-6pt}
\centering
\includegraphics[width=\linewidth]{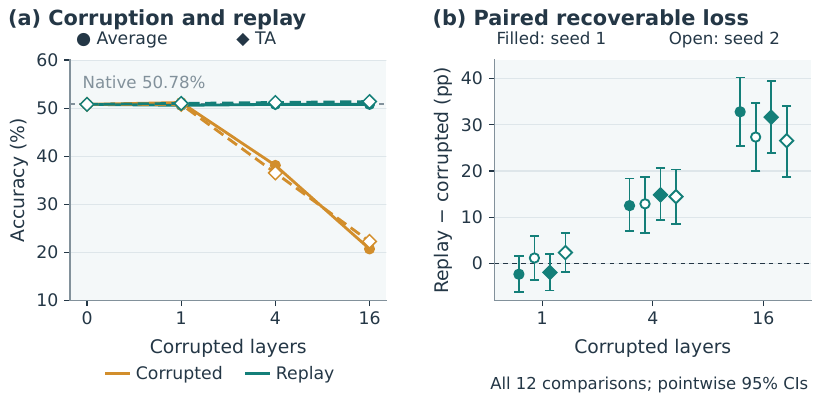}
\vspace{-2em}
\caption{\textbf{Task effects of controlled routing interventions.}
(a) Corrupted vs.\ clean-route replay accuracy.
(b) Paired replay gains with 95\% bootstrap CIs across all parent--permutation settings.}
\label{fig:task-grounded-tests}
\vspace{-8pt}
\end{wrapfigure}

\textbf{Known routing perturbations.}
We permute expert logits in nested sets of $k\in\{1,4,16\}$ OLMoE
layers under two fixed randomizations per Average and Task Arithmetic
parent. On 256 fixed ARC items per parent, we compare corrupted execution
with replay of that parent's clean route schedule.
Figure~\ref{fig:task-grounded-tests}(a) shows the accuracy changes;
panel (b) retains all twelve paired effects and their uncertainty.
Four-layer perturbations yield $12.50$--$14.84$ percentage points (pp)
of recovery, and full-layer perturbations yield $26.56$--$32.81$ pp.
All eight comparisons at $k=4,16$ pass Holm correction across the twelve
tests; none at $k=1$ does. One-layer perturbations do not consistently
lower accuracy.
Replay accuracy ranges from $50.39\%$ to $51.56\%$ around the $50.78\%$
clean reference, so this is not reconstruction of the clean forward
pass. These controls establish recovery of the tested imposed losses,
not a repair learned by SRR.

\textbf{Natural routing alternatives.}
A natural reference need not be appropriate for the merged experts.
Table~\ref{tab:natural-routing-tests} brings together the science-domain
source replays from Section~\ref{sec:route-diagnosis} and separate OLMoE
replays of a frozen LC-calibrated route schedule.
This intervention replays source-derived routing decisions computed from
the source forward; it is not a source-router parameter swap.
A swapped source router would instead recompute routing on hidden states
produced by the intervened merged network, and therefore defines a
different counterfactual policy.
Neither accuracy nor correct-choice margin---the correct option's score
minus the highest incorrect-option score---has a strictly positive
interval in any displayed setting.
LC replay changes correctness on just one item per parent, yielding
$-0.39$ pp for Average and $+0.39$ pp for TA.
These results do not establish recovery under the tested natural references.
Thus, routing failure requires evidence of recoverable task loss under a
specified intervention, not source disagreement alone.
We next use source information to construct selective router updates and
evaluate their effects separately from their fitting objective.

\section{Selective Router Repair}
\label{sec:srr}

\begin{wraptable}{r}{0.50\columnwidth}
\vspace{-2.5em}
\centering
\caption{\textbf{Natural routing alternatives.}
Changes from native routing with paired 95\% bootstrap CIs.
$m$ denotes the correct-choice margin.}
\label{tab:natural-routing-tests}
\begingroup
\scriptsize
\setlength{\tabcolsep}{1.8pt}
\renewcommand{\arraystretch}{1.06}
\arrayrulecolor{RTRule}
\resizebox{\linewidth}{!}{%
\begin{tabular}{@{\hspace{2pt}}lccc@{\hspace{2pt}}}
\toprule[0.65pt]
\rowcolor{RTHeader}
\textbf{Model / merge} & $N$ & $\Delta\mathrm{Acc}$ (pp) & $\Delta m$ ($10^{-3}$) \\
\midrule[0.35pt]
\rowcolor{RTBlue}
\multicolumn{4}{@{\hspace{2pt}}l@{}}{\strut\textbf{Source-route replay: science}} \\
DeepSeekMoE / Avg. & 256 & $0.00\,[-1.56,+1.56]$ & $-5.2\,[-18.2,+7.9]$ \\
DeepSeekMoE / TA & 256 & $+0.78\,[-0.78,+2.34]$ & $+2.1\,[-11.8,+17.0]$ \\
OLMoE / Avg. & 256 & $+1.17\,[-1.56,+3.91]$ & $+2.5\,[-26.1,+31.2]$ \\
OLMoE / TA & 256 & $+1.17\,[-1.17,+3.52]$ & $-6.2\,[-41.8,+28.5]$ \\
Qwen3-MoE / Avg. & 64 & $-1.56\,[-4.69,0.00]$ & $+8.8\,[-12.0,+28.2]$ \\
Qwen3-MoE / TA & 64 & $-1.56\,[-4.69,0.00]$ & $+1.8\,[-16.9,+19.7]$ \\
\addlinespace[1pt]
\rowcolor{RTSand}
\multicolumn{4}{@{\hspace{2pt}}l@{}}{\strut\textbf{LC replay: ARC}} \\
OLMoE / Avg. & 256 & $-0.39\,[-1.17,0.00]$ & $-0.3\,[-3.7,+3.2]$ \\
OLMoE / TA & 256 & $+0.39\,[0.00,+1.17]$ & $-2.6\,[-6.6,+0.8]$ \\
\bottomrule[0.65pt]
\end{tabular}%
}
\endgroup
\vspace{-3em}
\end{wraptable}

Section~\ref{sec:task-grounded-failure} separates candidate interventions
from evidence of recovery. We construct \textbf{Selective Router Repair
(SRR)} to study selective source supervision, then test its local directions,
executed corrections, and task effects.

\subsection{Selecting Expert Pairs}
\label{sec:srr-construction}
\label{sec:srr-pair-selection-main}

Let $B$, $S_d$, and $M$ denote the shared base, domain-$d$ specialist,
and merged parent. On identical parent-generated continuations, each model
uses its own hidden states. For response token $y_t$ after prefix $c_t$,
write $n_t^A=-\log P_A(y_t\mid c_t)$ and let $z_t^A\in\mathbb R^E$
be the router logits over $E$ experts. Omitting layer indices and writing
$S=S_{d_t}$, we form a source--base preference profile for each prompt $i$:
\begin{equation}
\bar u_i=
\frac{\sum_{t\in i}\omega_t\operatorname{center}(z_t^S-z_t^B)}
     {\sum_{t\in i}\omega_t},
\qquad
\omega_t=\sigma\!\left((n_t^B-n_t^S)/\tau\right),
\label{eq:srr-main-profile}
\end{equation}
where $\sigma$ is the sigmoid and $\operatorname{center}$ subtracts the
expert-wise mean. Activity is the similarly weighted maximum of source
and base routing probabilities. We average both quantities equally over
prompts in each domain and construction split.
Pairs join active experts with opposite, consistent profile signs across
both splits. Ranking combines profile separation, sign consistency, and
activity; selection cycles across domains without reusing an expert. 
The fixed screen requires activity $\geq0.002$ and sign consistency
$\geq0.75$, selecting two pairs per domain and layer without benchmark scores.

\subsection{Fitting Router Corrections}
\label{sec:srr-fitting-main}

For pair $j=(a_j,b_j)$ assigned to domain $d_j$, let
$m_{tj}^A=z_{t,a_j}^A-z_{t,b_j}^A$ and $r_{tj}=m_{tj}^S-m_{tj}^M$.
We clip this source--parent residual by the source--base difference and
weight matching-domain tokens on which the specialist outpredicts the parent:
\begin{equation}
\begin{aligned}
\widetilde r_{tj}
&=\operatorname{sgn}(r_{tj})
\min\!\left\{|r_{tj}|,\,|m_{tj}^S-m_{tj}^B|\right\},\\
\rho_{tj}
&=\mathbf1[d_t=d_j]\mathbf1[\delta_t>0]\sigma(\delta_t/\tau),
\qquad \delta_t=n_t^M-n_t^S.
\end{aligned}
\label{eq:srr-main-target}
\end{equation}
Unlike $\omega_t$, which measures source advantage over the base,
$\rho_{tj}$ compares the source with the parent. Neither weight is a
label of routing failure.
For cached parent inputs $h_t^M\in\mathbb R^{d_h}$ and $N$ training tokens,
we solve
\begin{equation}
\begin{aligned}
v_j^\star
&=\arg\min_v\;\frac12\sum_{t=1}^N\rho_{tj}
\big((h_t^M)^\top v-\widetilde r_{tj}\big)^2
+\frac{\lambda_j}{2}\|v\|_2^2,\\
\lambda_j
&=\gamma\max\!\left\{N^{-1}\sum_{t=1}^N\rho_{tj}\|h_t^M\|_2^2,
10^{-12}\right\}.
\end{aligned}
\label{eq:srr-main-fit}
\end{equation}
Writing $w_e^\top$ for router row $e$, the update is
\begin{equation}
w'_{a_j}=w^M_{a_j}+\frac{\eta}{2}v_j^\star,
\qquad
w'_{b_j}=w^M_{b_j}-\frac{\eta}{2}v_j^\star.
\label{eq:srr-main-update}
\end{equation}
This changes the pair gap by $\eta h^\top v_j^\star$ on fixed input $h$.
Only selected rows in the final five sparse layers change; all fits precede
joint writeback, without refreshing traces. We use $\tau=0.5$,
$\gamma=10^{-3}$, $\eta=0.0625$, and preconditioned conjugate gradients
(Appendix~\ref{app:srr-implementation}).
The objective fits selected logit residuals rather than task loss.
A two-expert counterexample shows why better proxy fit need not improve
token likelihood (Appendix~\ref{app:srr-counterexample}).

\subsection{Local Supervision and Executed Corrections}
\label{sec:supervision-utility}

\textbf{Do source predictions identify useful directions?}
We evaluate 256 independent diagnostic prompts for each OLMoE and Qwen3-MoE
under Average and TA candidate, selecting one token--layer--pair event per prompt.
At event $e$, equal-magnitude positive and opposite logit perturbations
follow either the source-informed or fitted direction $q$:
\begin{equation}
U_e^q=\ell_e^{\mathrm{native}}-\ell_e^{q,+},
D_e^q=\ell_e^{q,-}-\ell_e^{q,+},
\label{eq:srr-local-contrasts}
\end{equation}
\begin{wrapfigure}{r}{0.55\columnwidth}
\vspace{-1.9em}
\centering
\includegraphics[width=\linewidth]{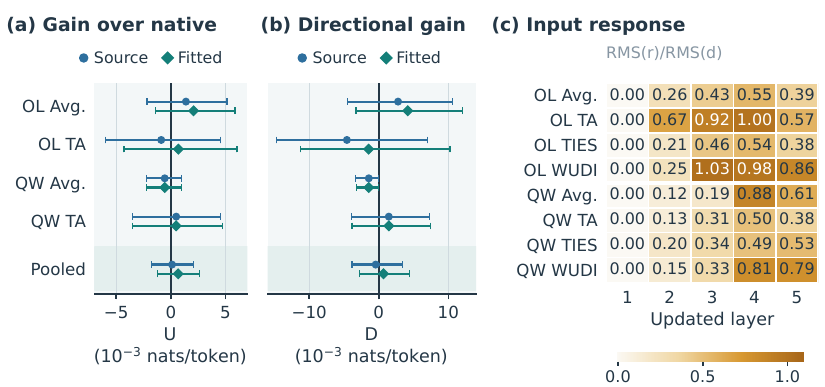}
\vspace{-2.3em}
\caption{\textbf{Local utility and input response of SRR.}
(a,b) Positive-support NLL gains over native ($U$) and opposite ($D$) routing, with 95\% bootstrap CIs.
(c) Input-response/direct RMS ratios across eight settings and five updated layers.}
\label{fig:srr-local-diagnostics}
\vspace{-1em}
\end{wrapfigure}
where $\ell_e$ is next-token NLL. Positive $U$ improves over native;
positive $D$ favors the proposed direction over its opposite.
Of 1024 events, 471 have positive source advantage.
Source and fitted directions agree in sign on $83.4\%$ of supported events
(95\% CI $[80.0,86.6]\%$), yet neither pooled $U$ nor $D$ establishes
an improvement (Figure~\ref{fig:srr-local-diagnostics}(a,b), OL and QW denote OLMoE and Qwen3-MoE).
The matched source-direction comparison also does not establish
enrichment (Appendix~\ref{app:source-interventions}).
Directional agreement therefore does not supply the missing evidence of
local utility.

\textbf{Do fitted corrections execute as predicted?}
All fits use parent inputs, but earlier updates can change later inputs.
On identical continuations through eight OLMoE/Qwen3-MoE Parent/SRR pairs,
we measure
one event per prompt in each updated layer: 128 prompts per setting and
5,120 events in total. At reference precision, the selected pair's logit change decomposes as
\begin{equation}
\Delta m_{t,ab}=
\underbrace{(\Delta w_a-\Delta w_b)^\top h_t^M}_{d_{t,ab}:\ \text{direct correction}}
+
\underbrace{(w'_a-w'_b)^\top(h_t^{\mathrm{SRR}}-h_t^M)}_{r_{t,ab}:\ \text{input response}}.
\label{eq:srr-main-execution}
\end{equation}
The input response is zero at the first updated layer. At the final layer,
its reported RMS is approximately $0.38$--$0.86$ times the direct-correction
RMS across settings (Figure~\ref{fig:srr-local-diagnostics}(c)).
Thus, joint execution need not reproduce the fixed-input correction.
The figure summarizes the reference-precision decomposition; native-forward
discrepancies, absolute RMS intervals, and routing-set changes are reported
in Appendix~\ref{app:srr-execution}. This decomposition does not attribute
task effects to input response.

\section{Experiments}
\label{sec:experiments}
\label{sec:checkpoint-effects}

Section~\ref{sec:srr} examines the construction and execution of SRR.
We now test whether these updates recover downstream performance under
Section~\ref{sec:task-grounded-failure}'s task-grounded criterion.

\subsection{Experimental Setup}
\label{sec:experimental-setup}

\begin{table*}[t]
  \centering
  \caption{\textbf{Task scores across three merged MoEs.}
  Avg. equally weights eight benchmarks; $\Delta$Avg. compares each method with its own Parent (pp).
  The 95\% CI (pp) is the paired item-bootstrap 95\% confidence interval for the SRR--Parent difference in percentage points.
  HARC scores use the same matched requests.}
  \label{tab:main-results}
  \begingroup
  \SRRFullTableFont
  \setlength{\tabcolsep}{1.6pt}
  \renewcommand{\arraystretch}{1.08}
  \arrayrulecolor{SRRFullRule}
  \resizebox{\textwidth}{!}{
  \begin{tabular}{@{\hspace{3pt}}l*{12}{r}@{\hspace{3pt}}}
    \toprule[0.65pt]
    \rowcolor{SRRFullHeader}
    & \multicolumn{8}{c}{\textbf{Benchmarks (\%)}}
    & \multicolumn{2}{c}{\textbf{Summary}}
    & \multicolumn{2}{c}{\textbf{95\% CI (pp)}} \\
    \cmidrule(lr){2-9}\cmidrule(lr){10-11}\cmidrule(lr){12-13}
    \rowcolor{SRRFullHeader}
    \textbf{Method} & \textbf{MMLU}
    & \shortstack{\textbf{Hella}\\\textbf{Swag}}
    & \textbf{ARC-C} & \textbf{ARC-E} & \textbf{PIQA}
    & \shortstack{\textbf{Wino}\\\textbf{Grande}}
    & \textbf{BoolQ} & \textbf{GSM8K}
    & \textbf{Avg.} & \textbf{$\Delta$Avg.}
    & \textbf{Lower} & \textbf{Upper} \\
    \midrule[0.35pt]
    \rowcolor{SRRFullBand}
    \multicolumn{13}{@{\hspace{3pt}}l@{\hspace{3pt}}}{\strut\color{SRRFullInk}\textbf{DeepSeekMoE-16B-A3B}} \\
    Average & 57.60 & 80.02 & 57.68 & 83.50 & 82.48 & 76.09 & 81.74 & 48.67 & 70.97 & \textemdash & \textemdash & \textemdash \\
    \rowcolor{SRRFullHARCTint}
    \hspace{0.65em}{\color{SRRFullHARCInk}\textbf{+ HARC}} & 57.58 & 80.05 & 57.68 & 84.01 & 82.32 & 76.48 & 82.05 & 49.51 & 71.21 & +0.236 & \textemdash & \textemdash \\
    \rowcolor{SRRFullSRRTint}
    \hspace{0.65em}{\color{SRRFullTeal}\textbf{+ SRR}} & 57.68 & 79.99 & 57.68 & 83.50 & 82.48 & 76.09 & 81.80 & 48.67 & 70.99 & +0.015 & -0.011 & +0.041 \\
    \addlinespace[1pt]
    Task Arithmetic & 58.04 & 80.03 & 58.11 & 83.92 & 82.32 & 76.48 & 80.83 & 45.26 & 70.62 & \textemdash & \textemdash & \textemdash \\
    \rowcolor{SRRFullHARCTint}
    \hspace{0.65em}{\color{SRRFullHARCInk}\textbf{+ HARC}} & 58.07 & 80.05 & 57.94 & 84.01 & 81.94 & 76.72 & 80.61 & 45.56 & 70.61 & -0.012 & \textemdash & \textemdash \\
    \rowcolor{SRRFullSRRTint}
    \hspace{0.65em}{\color{SRRFullTeal}\textbf{+ SRR}} & 58.04 & 80.03 & 58.11 & 83.92 & 82.32 & 76.48 & 80.83 & 45.26 & 70.62 & +0.000 & +0.000 & +0.000 \\
    \addlinespace[1pt]
    TIES & 56.50 & 80.34 & 57.76 & 82.03 & 82.26 & 76.01 & 82.72 & 54.81 & 71.56 & \textemdash & \textemdash & \textemdash \\
    \rowcolor{SRRFullHARCTint}
    \hspace{0.65em}{\color{SRRFullHARCInk}\textbf{+ HARC}} & 56.42 & 80.46 & 56.91 & 82.58 & 82.43 & 76.48 & 82.51 & 53.60 & 71.42 & -0.133 & \textemdash & \textemdash \\
    \rowcolor{SRRFullSRRTint}
    \hspace{0.65em}{\color{SRRFullTeal}\textbf{+ SRR}} & 56.43 & 80.33 & 57.76 & 82.03 & 82.15 & 76.01 & 82.72 & 54.81 & 71.53 & -0.024 & -0.054 & +0.004 \\
    \addlinespace[1pt]
    WUDI-Merge & 56.88 & 79.47 & 55.12 & 81.36 & 82.15 & 77.03 & 82.29 & 52.39 & 70.84 & \textemdash & \textemdash & \textemdash \\
    \rowcolor{SRRFullHARCTint}
    \hspace{0.65em}{\color{SRRFullHARCInk}\textbf{+ HARC}} & 56.78 & 79.41 & 55.12 & 81.23 & 82.26 & 77.11 & 82.17 & 52.92 & 70.87 & +0.039 & \textemdash & \textemdash \\
    \rowcolor{SRRFullSRRTint}
    \hspace{0.65em}{\color{SRRFullTeal}\textbf{+ SRR}} & 56.97 & 79.51 & 55.17 & 81.43 & 82.21 & 77.03 & 82.32 & 52.51 & 70.89 & +0.057 & -0.025 & +0.059 \\
    \midrule[0.4pt]
    \rowcolor{SRRFullBand}
    \multicolumn{13}{@{\hspace{3pt}}l@{\hspace{3pt}}}{\strut\color{SRRFullInk}\textbf{OLMoE-7B-A1B}} \\
    Average & 55.47 & 80.15 & 57.08 & 81.52 & 81.83 & 73.72 & 80.40 & 58.15 & 71.04 & \textemdash & \textemdash & \textemdash \\
    \rowcolor{SRRFullHARCTint}
    \hspace{0.65em}{\color{SRRFullHARCInk}\textbf{+ HARC}} & 55.56 & 80.18 & 57.25 & 81.44 & 81.61 & 74.11 & 80.52 & 58.98 & 71.21 & +0.168 & \textemdash & \textemdash \\
    \rowcolor{SRRFullSRRTint}
    \hspace{0.65em}{\color{SRRFullTeal}\textbf{+ SRR}} & 55.47 & 80.15 & 57.17 & 81.44 & 81.83 & 73.09 & 80.40 & 59.89 & 71.18 & +0.139 & -0.063 & +0.336 \\
    \addlinespace[1pt]
    Task Arithmetic & 55.45 & 79.78 & 57.59 & 82.32 & 81.34 & 72.93 & 79.36 & 56.79 & 70.69 & \textemdash & \textemdash & \textemdash \\
    \rowcolor{SRRFullHARCTint}
    \hspace{0.65em}{\color{SRRFullHARCInk}\textbf{+ HARC}} & 55.41 & 79.84 & 57.68 & 82.37 & 81.39 & 72.14 & 79.57 & 57.54 & 70.74 & +0.048 & \textemdash & \textemdash \\
    \rowcolor{SRRFullSRRTint}
    \hspace{0.65em}{\color{SRRFullTeal}\textbf{+ SRR}} & 55.62 & 79.93 & 57.52 & 82.42 & 81.38 & 73.11 & 79.43 & 57.11 & 70.82 & +0.121 & -0.163 & +0.208 \\
    \addlinespace[1pt]
    TIES & 54.37 & 80.19 & 56.14 & 80.01 & 81.34 & 71.82 & 81.07 & 55.50 & 70.06 & \textemdash & \textemdash & \textemdash \\
    \rowcolor{SRRFullHARCTint}
    \hspace{0.65em}{\color{SRRFullHARCInk}\textbf{+ HARC}} & 54.56 & 80.13 & 56.66 & 79.80 & 81.39 & 71.59 & 81.28 & 55.42 & 70.10 & +0.048 & \textemdash & \textemdash \\
    \rowcolor{SRRFullSRRTint}
    \hspace{0.65em}{\color{SRRFullTeal}\textbf{+ SRR}} & 54.49 & 80.29 & 55.99 & 79.98 & 81.49 & 72.24 & 81.14 & 55.60 & 70.15 & +0.096 & -0.230 & +0.222 \\
    \addlinespace[1pt]
    WUDI-Merge & 54.12 & 79.52 & 52.39 & 76.26 & 81.23 & 72.53 & 81.99 & 56.10 & 69.27 & \textemdash & \textemdash & \textemdash \\
    \rowcolor{SRRFullHARCTint}
    \hspace{0.65em}{\color{SRRFullHARCInk}\textbf{+ HARC}} & 54.19 & 79.56 & 52.73 & 76.39 & 80.96 & 72.53 & 81.62 & 55.57 & 69.19 & -0.075 & \textemdash & \textemdash \\
    \rowcolor{SRRFullSRRTint}
    \hspace{0.65em}{\color{SRRFullTeal}\textbf{+ SRR}} & 54.13 & 79.51 & 52.47 & 76.35 & 81.23 & 72.85 & 82.02 & 55.42 & 69.25 & -0.021 & -0.233 & +0.188 \\
    \midrule[0.4pt]
    \rowcolor{SRRFullBand}
    \multicolumn{13}{@{\hspace{3pt}}l@{\hspace{3pt}}}{\strut\color{SRRFullInk}\textbf{Qwen3-30B-A3B}} \\
    Average & 81.96 & 83.14 & 66.72 & 89.18 & 82.43 & 75.93 & 87.46 & 90.30 & 82.14 & \textemdash & \textemdash & \textemdash \\
    \rowcolor{SRRFullHARCTint}
    \hspace{0.65em}{\color{SRRFullHARCInk}\textbf{+ HARC}} & 81.90 & 83.01 & 67.32 & 89.18 & 82.05 & 76.24 & 87.52 & 90.37 & 82.20 & +0.060 & \textemdash & \textemdash \\
    \rowcolor{SRRFullSRRTint}
    \hspace{0.65em}{\color{SRRFullTeal}\textbf{+ SRR}} & 81.88 & 83.13 & 66.47 & 89.10 & 82.32 & 76.40 & 87.37 & 90.14 & 82.10 & -0.039 & -0.212 & +0.130 \\
    \addlinespace[1pt]
    Task Arithmetic & 81.77 & 83.19 & 67.32 & 89.35 & 83.08 & 76.87 & 88.10 & 87.72 & 82.18 & \textemdash & \textemdash & \textemdash \\
    \rowcolor{SRRFullHARCTint}
    \hspace{0.65em}{\color{SRRFullHARCInk}\textbf{+ HARC}} & 81.80 & 83.13 & 67.58 & 89.52 & 82.81 & 77.66 & 87.86 & 87.72 & 82.26 & +0.084 & \textemdash & \textemdash \\
    \rowcolor{SRRFullSRRTint}
    \hspace{0.65em}{\color{SRRFullTeal}\textbf{+ SRR}} & 81.70 & 83.11 & 67.32 & 89.44 & 82.86 & 76.87 & 88.13 & 88.10 & 82.19 & +0.016 & -0.155 & +0.184 \\
    \addlinespace[1pt]
    TIES & 80.95 & 82.50 & 67.24 & 88.34 & 82.26 & 75.53 & 87.86 & 89.84 & 81.82 & \textemdash & \textemdash & \textemdash \\
    \rowcolor{SRRFullHARCTint}
    \hspace{0.65em}{\color{SRRFullHARCInk}\textbf{+ HARC}} & 81.58 & 82.61 & 67.75 & 88.01 & 82.21 & 75.06 & 87.37 & 88.86 & 81.68 & -0.136 & \textemdash & \textemdash \\
    \rowcolor{SRRFullSRRTint}
    \hspace{0.65em}{\color{SRRFullTeal}\textbf{+ SRR}} & 81.05 & 82.45 & 67.49 & 88.09 & 82.32 & 75.53 & 87.86 & 89.84 & 81.83 & +0.013 & -0.174 & +0.194 \\
    \addlinespace[1pt]
    WUDI-Merge & 81.83 & 83.49 & 62.54 & 86.03 & 81.99 & 74.51 & 87.09 & 87.79 & 80.66 & \textemdash & \textemdash & \textemdash \\
    \rowcolor{SRRFullHARCTint}
    \hspace{0.65em}{\color{SRRFullHARCInk}\textbf{+ HARC}} & 81.76 & 83.53 & 62.80 & 85.86 & 81.72 & 74.98 & 87.00 & 87.49 & 80.64 & -0.016 & \textemdash & \textemdash \\
    \rowcolor{SRRFullSRRTint}
    \hspace{0.65em}{\color{SRRFullTeal}\textbf{+ SRR}} & 81.91 & 83.52 & 62.71 & 85.90 & 81.61 & 74.51 & 87.09 & 87.57 & 80.60 & -0.056 & -0.261 & +0.150 \\
    \bottomrule[0.65pt]
  \end{tabular}
  }
  \endgroup
  \vspace{-1.8em}
\end{table*}

\textbf{Baselines:} We study DeepSeekMoE-16B-A3B \citep{dai2024deepseekmoe},
OLMoE-7B-A1B \citep{muennighoff2025olmoe}, and
Qwen3-30B-A3B \citep{yang2025qwen3}, compared with HARC \citep{huang2026model}.

\textbf{Merging Settings:} For each architecture, we evaluate Average merge \citep{ilharco2022editing}, Task Arithmetic (TA)
\citep{ilharco2022editing}, TIES \citep{yadav2023ties}, and state-of-the-art WUDI-Merge \citep{cheng2025whoever}.
These methods construct the merged parents; SRR is then applied to each parent.
Following Sec.~\ref{sec:srr}, it fits corrections on fixed parent-generated
continuations and updates selected router parameters in the final five sparse
layers, leaving all other parameters unchanged.

\textbf{Benchmarks and uncertainty.}
We evaluate MMLU \citep{hendrycks2020measuring}, HellaSwag \citep{zellers2019hellaswag}, 
ARC-Challenge (ARC-C), ARC-Easy (ARC-E) \citep{allenai:arc}, PIQA \citep{Bisk2020},
WinoGrande \citep{sakaguchi2021winogrande}, BoolQ \citep{clark2019boolq}, and GSM8K \citep{cobbe2021gsm8k}
with five-shot lm-evaluation-harness and a VLLM backend.
Each pair shares 35,326 item identities and is evaluated five times under the
fixed deterministic protocol. We retain each task's accuracy,
normalized-accuracy, or exact-match scoring rule.
Avg. is the equally weighted mean of the eight benchmark percentages;
$\Delta$Avg. is SRR minus its own parent, in percentage points (pp).
Paired item-bootstrap intervals resample matched records within tasks and
recompute the task aggregate. They quantify item uncertainty conditional
on fixed checkpoints, not variation across repair seeds.
Full scoring and checkpoint details appear in
Appendix~\ref{app:paired-evaluation}.

\textbf{Evaluation Settings:} All performance scores are averaged over five evaluation runs.
Scores are percentages, and \emph{Avg.} is their mean score.
We report \emph{$\Delta$Avg.} relative to the corresponding unrepaired parent
in percentage points (pp), and average these changes equally across the
12 settings for the overall result.
Base-model scores provide context but are not the repair baseline.

\textbf{Implementations:} All experiments are conducted on 8 A800 GPUs with same configuration and 
software environment to ensure consistency and reproducibility.

\subsection{Main Results}
\label{sec:main-results}

\textbf{Average effects across settings.}
Table~\ref{tab:main-results} compares HARC and SRR with their corresponding
merged parents. Among the eight OLMoE/Qwen3-MoE settings, five
average-score changes are positive, ranging overall from
$-0.056$ to $+0.139$ percentage points (pp).
OLMoE has positive point estimates under Average, TA, and TIES, whereas
Qwen3-MoE has positive estimates under TA and TIES. WUDI-Merge yields
negative estimates in both.

\textbf{Comparison with HARC.}
Across the eight active SRR settings, each method has the higher point
estimate in four cases. HARC leads under Average merging and SRR under
TIES on both architectures, while the ordering under TA and WUDI depends
on the backbone. Relative ranking need not imply recovery: on
OLMoE--WUDI, SRR exceeds HARC ($-0.021$ versus $-0.075$ pp), yet both fall
below Parent. On DeepSeekMoE, HARC likewise ranges from $-0.133$ pp under
TIES to $+0.236$ pp under Average merging.

\textbf{Task-level heterogeneity.}
Similar averages can conceal different task responses. On OLMoE--Average,
SRR gains $1.74$ pp on GSM8K but loses $0.63$ pp on WinoGrande; HARC gains
$0.83$ and $0.39$ pp, respectively. On Qwen3--TA, SRR improves GSM8K by
$0.38$ pp but decreases PIQA by $0.22$ pp. These differences do not
establish task-level significance. The cross-backbone comparison also
changes architecture and parameter count together, so it does not isolate
a model-size effect.
\begin{wraptable}{r}{0.45\textwidth}
  \vspace{-1.8em}
  \centering
  \footnotesize
  \setlength{\tabcolsep}{3pt}
  \renewcommand{\arraystretch}{1.15}
  \caption{\textbf{Matched selection effects.}
  Positive-minus-control source-direction gains
  ($10^{-3}$ nats/token; 440 matched pairs).}
  \label{tab:matched-selection-main}
  \arrayrulecolor{RTRule}
  \begin{tabularx}{\linewidth}{@{}lr>{\raggedleft\arraybackslash}X@{}}
    \toprule
    \rowcolor{RTHeader}
    Contrast & Estimate & 95\% CI \\
    \midrule
    Native ($U$) & $-0.341$ & $[-3.323,\,2.631]$ \\
    Opposite ($D$) & $-0.997$ & $[-5.653,\,3.868]$ \\
    \bottomrule
  \end{tabularx}
  \arrayrulecolor{black}
  \vspace{-1em}
\end{wraptable}
\textbf{Incremental value of likelihood selection.}
Task scores evaluate complete updates, not the contribution of their
selection rule. We therefore examine the independent local diagnostic
from Section~\ref{sec:supervision-utility}.
Positive-support events are matched with replacement to non-positive
controls within the same architecture, merge, domain, layer, and expert
pair, choosing the nearest native NLL.
Across 440 matched pairs, neither source-direction contrast establishes
higher utility on positive support (Table~\ref{tab:matched-selection-main}).
Unlike Figure~\ref{fig:srr-local-diagnostics}'s positive-support means,
these differences test whether selection adds useful local information.
They do not establish that the groups are equivalent or explain the
benchmark changes; matching and additional weighting diagnostics appear
in Appendix~\ref{app:local-matched-support}.

\textbf{Implications for routing diagnosis.}
These fitted updates differ from the source-route replay tested in
Section~\ref{sec:route-diagnosis}. Positive point estimates therefore do not
contradict the finding that source disagreement alone is insufficient
evidence of failure. Following Section~\ref{sec:task-grounded-failure},
recovery must be established for the specific intervention and task;
neither fitting source-derived targets nor outperforming another repair
method substitutes for evidence of improvement over the unrepaired parent.

\section{Conclusion}
\label{sec:conclusion}

In this paper, we show that \textbf{routing drift alone is insufficient evidence of routing
failure} in merged MoE LLMs.
Across DeepSeekMoE, OLMoE, and Qwen3-MoE, local interventions attribute most reassignments
to router-input shifts, while structural differences poorly predict
source-route intervention gains.
Morivated by this, we formalize intervention-relative recoverable task loss and develop
an \emph{analysis toolkit} for paired routing tests with fixed non-routing
parameters.
These tests detect recoverable loss under controlled router corruption.
To examine source-informed supervision, we propose \emph{Selective Router
Repair (SRR)}, which fits likelihood-weighted expert-pair corrections on merged
router inputs.
Its evaluation finds no reliable evidence that source-likelihood advantages
identify beneficial local corrections or that fitted updates improve average
task performance on the tested checkpoints.
Together, these contributions separate candidate construction from demonstrated
recovery: task-level intervention effects, rather than source agreement,
must justify repair.

\subsection*{AI use statement}
We used generative AI tools to polish the manuscript and figures, help test experiments, and check code and infrastructure.
The authors reviewed the AI-assisted work, verified the reported results and claims, and take full responsibility for the submission.

\subsection*{Reproducibility statement}
The anonymized code, toolkit, and configurations are available at \url{https://anonymous.4open.science/r/SRR-DDD2}.
Section~\ref{sec:experimental-setup} specifies the training settings and benchmarks; 
Appendix~\ref{app:paired-evaluation} document the configuration and validation procedure.

\bibliography{iclr2027_conference}
\bibliographystyle{iclr2027_conference}

\clearpage
\appendix
\setcounter{figure}{0}
\setcounter{table}{0}
\renewcommand{\thefigure}{\Alph{section}\arabic{figure}}
\renewcommand{\thetable}{\Alph{section}\arabic{table}}
\renewcommand{\theHfigure}{appendix.\Alph{section}.\arabic{figure}}
\renewcommand{\theHtable}{appendix.\Alph{section}.\arabic{table}}

\section{Routing Analysis: Protocols and Complete Results}
\label{app:representation-routing}

This appendix supports Section~\ref{sec:routing-drift}. We specify the
sampling and intervention conventions, then report the complete structural,
local-likelihood, and mixture-output comparisons. These measurements use
different event populations and are not pooled with the SRR task evaluation.

\subsection{Crossed interventions}
\label{app:representation-routing-design}

Let $\mathcal R(W,h)=\operatorname{Top\text{-}}k(Wh)$ denote the native
expert selection, with router parameters $W$ and input $h$. For source
and merged quantities, the four states are
\begin{equation}
\begin{aligned}
 r_{SS}&=\mathcal R(W_S,h_S), & r_{MS}&=\mathcal R(W_M,h_S),\\
 r_{SM}&=\mathcal R(W_S,h_M), & r_{MM}&=\mathcal R(W_M,h_M).
\end{aligned}
\label{eq:four-state-routing}
\end{equation}
The first subscript identifies the router and the second its input.
Conditioning on $r_{SS}\ne r_{MM}$, we call a change representation-induced
when $r_{SM}\ne r_{SS}$ but $r_{MS}=r_{SS}$; router-parameter-induced is
the converse. Both crossed routes changing the source set defines a
redundant case; neither changing it defines an interaction-only case.
These categories concern whether an isolated replacement changes the
source set, not whether it reproduces the merged set. They establish local
sufficiency under the crossed intervention, not the origin of every upstream
representation change.

\subsection{Aligned Representations under Merging and SRR}
\label{app:illustrative-representation-trajectories}

\begin{figure*}[tbp]
  \centering
  \includegraphics[width=\textwidth]{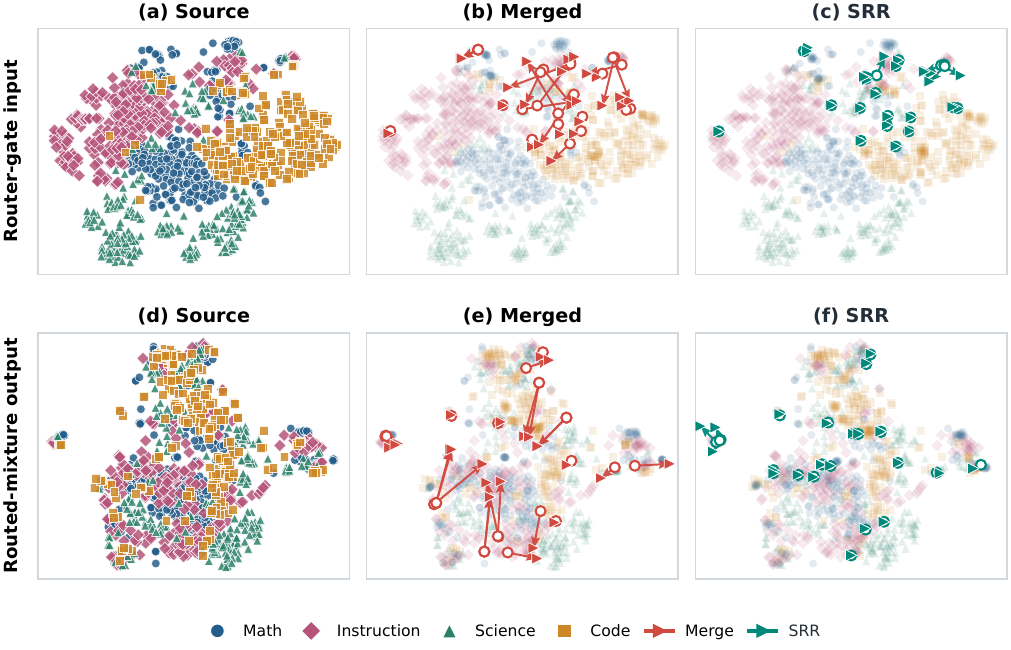}
  \caption{\textbf{Aligned representations in OLMoE.}
  Joint t-SNE at layer 15: router inputs (top) and routed-mixture outputs
  (bottom) under Source, Merged, and SRR. Red and teal arrows highlight
  displayed token displacements associated with merging and SRR, respectively.}
  \label{fig:appendix-expanded-representation}
\end{figure*}

\begin{figure*}[tbp]
  \centering
  \includegraphics[width=\textwidth]{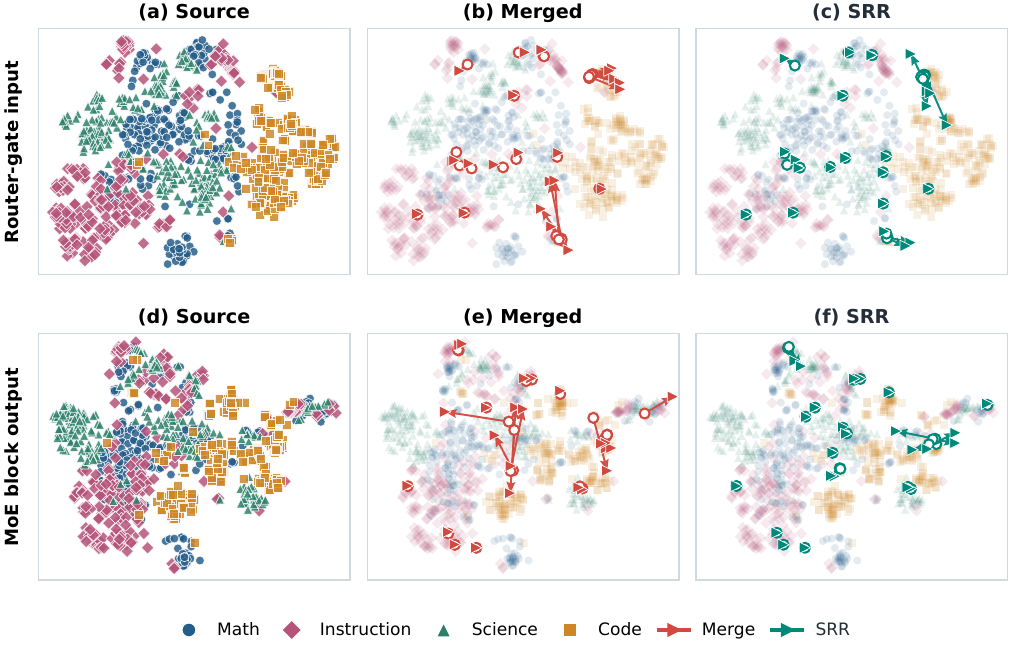}
  \caption{\textbf{Aligned representations in DeepSeekMoE.}
  Joint t-SNE of router inputs (top) and MoE-block outputs (bottom) at
  layer 27 under Source, Merged, and SRR. Display conventions follow
  Figure~\ref{fig:appendix-expanded-representation}.}
  \label{fig:appendix-deepseek-representation}
\end{figure*}

\begin{figure*}[tbp]
  \centering
  \includegraphics[width=\textwidth]{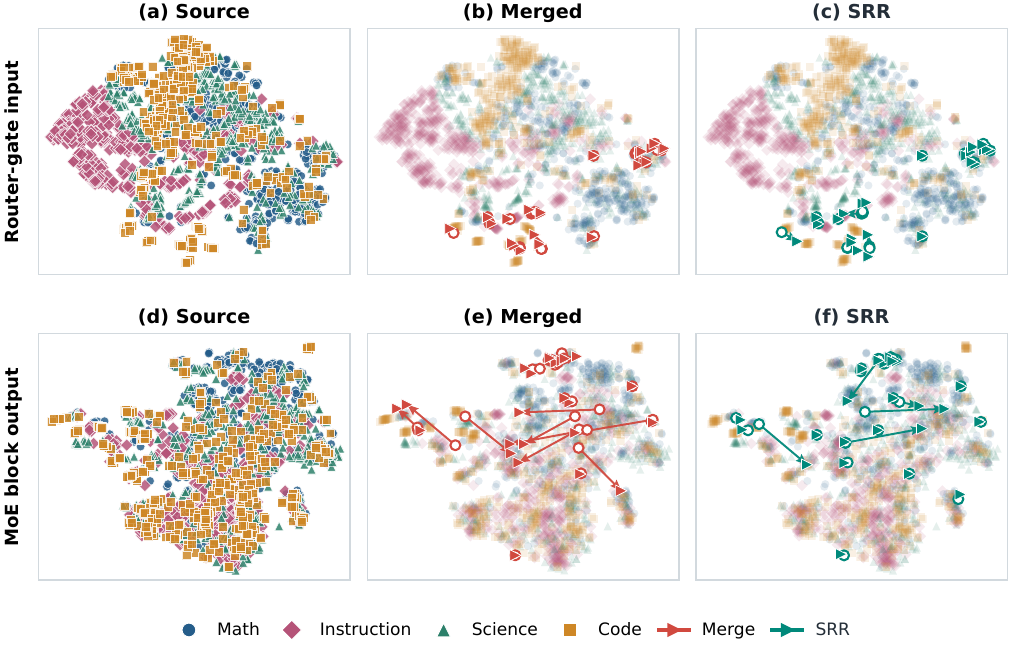}
  \caption{\textbf{Aligned representations in Qwen3-MoE.}
  Joint t-SNE of router inputs (top) and MoE-block outputs (bottom) at
  layer 47 under Source, Merged, and SRR. Display conventions follow
  Figure~\ref{fig:appendix-expanded-representation}.}
  \label{fig:appendix-qwen3-representation}
\end{figure*}

Figures~\ref{fig:appendix-expanded-representation}--\ref{fig:appendix-qwen3-representation}
use joint t-SNE to visualize aligned token representations from the source
specialist, merged parent, and SRR checkpoint. Each figure shows router
inputs above and layer outputs below: routed-mixture outputs for OLMoE,
and MoE-block outputs for DeepSeekMoE and Qwen3-MoE. Colors and shapes
identify domains, not experts. Red and teal arrows highlight token
displacements associated with merging and SRR, respectively.

\paragraph{OLMoE.}
At layer 15, domain-associated regions in the router-input projection
contrast with interleaved domain colors in the routed-mixture view
(Figure~\ref{fig:appendix-expanded-representation}). Merge and SRR arrows
show token-dependent changes at both measurement sites, extending the
input-only illustration in Figure~\ref{fig:representation-drift}.

\paragraph{DeepSeekMoE.}
At layer 27, the router-input and MoE-block-output views both contain
heterogeneous token shifts (Figure~\ref{fig:appendix-deepseek-representation}).
The block-output view retains visible domain overlap while highlighting
token-specific changes under SRR.

\paragraph{Qwen3-MoE.}
The layer-47 projections contain overlapping domain regions
(Figure~\ref{fig:appendix-qwen3-representation}). In the block-output view,
the highlighted SRR paths follow different directions, illustrating
heterogeneous changes within the projected layout.

\paragraph{Relation to the quantitative analyses.}
These views complement the input-shift analysis in Section~\ref{sec:route-origin}
and the execution analysis in Section~\ref{sec:supervision-utility}.
They compare model representations, unlike Section~\ref{sec:mixture-redundancy}'s
route comparisons at a fixed merged state with fixed experts.
Projected arrow lengths do not quantify original-space changes, and are
not compared across rows or architectures. Task recovery is evaluated
separately under Section~\ref{sec:task-grounded-failure}'s criterion.

\FloatBarrier

\subsection{Sampling and uncertainty}
\label{app:estimand-ledger}

Table~\ref{tab:estimand-ledger} records the original DeepSeekMoE/OLMoE
sampling frames. Qwen3-specific counts are given in
Appendix~\ref{app:qwen-route-diagnosis}. Prompts are balanced across
mathematics, instruction following, science, and code; hash selections do
not inspect intervention outcomes. Natural-frequency attribution includes
all continuation tokens and sparse layers, whereas local functional tests
concentrate on the final five sparse layers. The original natural-frequency
records contain 861,651 DeepSeekMoE and 510,656 OLMoE token--layer events.

\begin{table*}[t]
\centering
\footnotesize
\renewcommand{\arraystretch}{1.08}
\caption{Sampling frames for the original DeepSeekMoE/OLMoE analyses. Hash selections do not inspect outcomes. Qwen3-specific counts appear in Appendix~\ref{app:qwen-route-diagnosis}.}
\label{tab:estimand-ledger}
\begin{tabular}{@{}>{\raggedright\arraybackslash}p{0.20\textwidth}>{\raggedright\arraybackslash}p{0.47\textwidth}>{\raggedright\arraybackslash}p{0.25\textwidth}@{}}
\toprule
Analysis & Sampling frame & Record / inference unit \\
\midrule
Natural route origin & 256 prompts (64/domain); all continuation tokens and sparse layers & Token--layer event / prompt UUID \\
Structural diagnostic & Same 256 prompts; final five sparse layers; eight hash-selected events per prompt (2,048/cell) & Counterfactual event / prompt UUID \\
Token-local route patch & 80 changed-route events per architecture--parent cell & Counterfactual event / prompt UUID \\
Local reference and redundancy & One hash-selected event per prompt (256/cell); 112--149 changed-route events/cell & Prompt event / prompt UUID \\
Task-grounded route patch & 256 fixed items/domain/cell; every non-padding question--choice token and sparse layer & Multiple-choice item / item \\
Matched random routes & 32 matched alternatives for each changed-route event & Prompt event / prompt UUID \\
\bottomrule
\end{tabular}
\end{table*}

Uncertainty resamples prompts, retaining their dependent token--layer
records. Attribution uses 10,000 draws and recomputes record-weighted
ratios; metric associations use 4,000 draws with ranks recomputed each time.
Mixture comparisons first average within prompts and use 10,000 prompt
resamples. Task comparisons instead resample matched items 10,000 times.
Intervals are percentile intervals unless a separate multiplicity procedure
is specified. These choices preserve each estimand rather than treating
all tokens or layers as independent observations.

\subsection{Routing origin across architectures and domains}
\label{app:cross-architecture-attribution}

DeepSeekMoE uses Top-6 routing over 27 sparse layers; OLMoE uses Top-8 over
16. Table~\ref{tab:appendix-representation-router-attribution} reports
record-weighted attribution under Average and Task Arithmetic (TA).
The corresponding Qwen3 Top-8, 48-layer results are in
Table~\ref{tab:qwen-routing-origin}. Attribution conditions on observed
set changes, so these shares are distinct from the prevalence of changes.

\begin{table*}[t]
\centering
\small
\caption{Local four-state attribution. Values are percentages with 95\% prompt-bootstrap intervals; attribution shares condition on a changed source--merged expert set.}
\label{tab:appendix-representation-router-attribution}
\begin{tabular}{lccc}
\toprule
Architecture / Parent & Changed route (\%) & Representation (\%) & Router parameters (\%) \\
\midrule
DeepSeekMoE / Average & $33.9\,[32.2,35.7]$ & $77.7\,[77.1,78.2]$ & $1.79\,[1.65,1.94]$ \\
DeepSeekMoE / Task Arithmetic & $46.7\,[44.9,48.7]$ & $78.5\,[78.1,79.0]$ & $1.14\,[1.04,1.23]$ \\
OLMoE / Average & $42.7\,[41.3,44.1]$ & $85.8\,[85.4,86.1]$ & $0.62\,[0.57,0.68]$ \\
OLMoE / Task Arithmetic & $54.3\,[52.9,55.7]$ & $87.0\,[86.6,87.4]$ & $0.36\,[0.33,0.40]$ \\
\bottomrule
\end{tabular}
\end{table*}

\begin{figure*}[t]
  \centering
  \includegraphics[width=\textwidth]{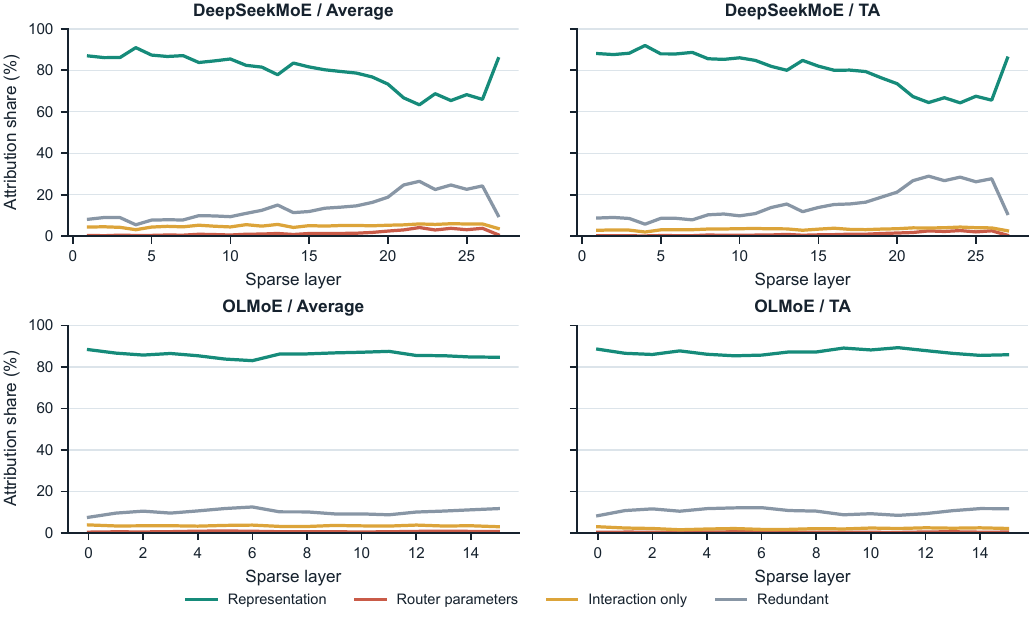}
  \caption{\textbf{Attribution across sparse layers.} Four-state shares among changed routes for DeepSeekMoE/OLMoE under Average and TA. Categories are defined in Appendix~\ref{app:representation-routing-design}.}
  \label{fig:appendix-layerwise-attribution}
\end{figure*}

\begin{figure*}[t]
  \centering
  \includegraphics[width=\textwidth]{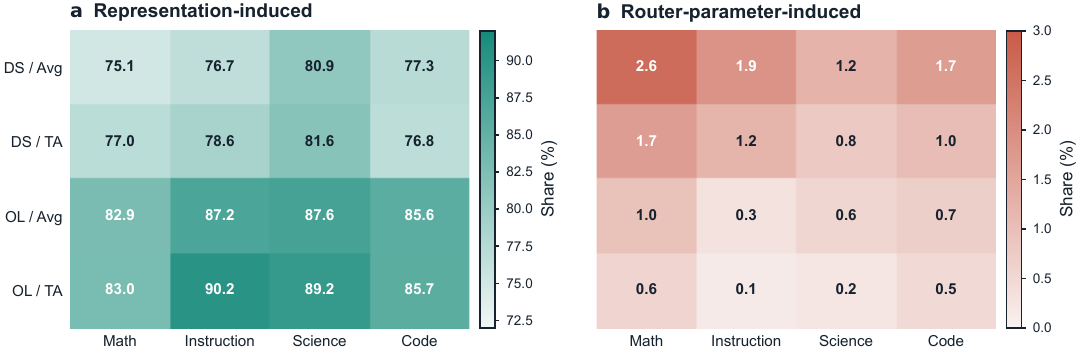}
  \caption{\textbf{Attribution by domain.} Representation-induced and router-parameter-induced shares use separate color scales; each cell reports its percentage.}
  \label{fig:appendix-domain-attribution}
\end{figure*}

Figures~\ref{fig:appendix-layerwise-attribution} and
\ref{fig:appendix-domain-attribution} retain the full depth and domain
profiles for DeepSeekMoE/OLMoE. Representation-induced changes dominate
throughout OLMoE and most DeepSeekMoE layers. Later DeepSeekMoE layers have
more redundant cases, but few router-parameter-only cases. Across domains,
representation-induced shares range from $75.1\%$ to $90.2\%$, versus
$0.1\%$ to $2.6\%$ for router-parameter-induced changes. The two domain
heatmaps use separate, explicitly labeled color scales.

\subsection{Across-layer representation association}
\label{app:layerwise-association}

\begin{figure*}[ht]
  \centering
  \includegraphics[width=0.92\textwidth]{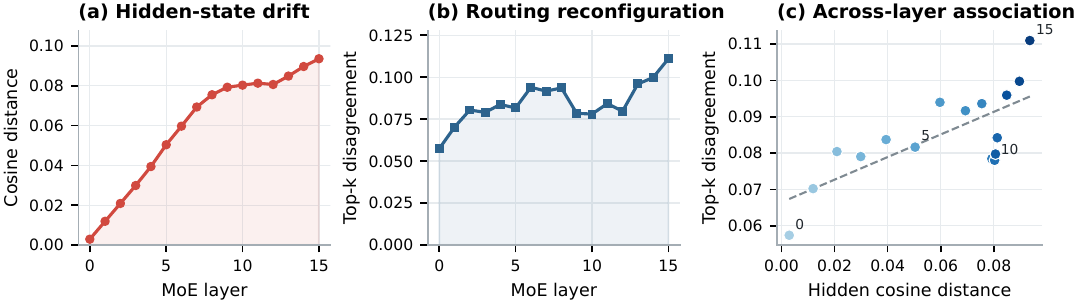}
  \caption{\textbf{Representation and routing across OLMoE depth.} Original-space hidden-state cosine distance and Top-$k$ disagreement co-vary across 16 layers. This association is distinct from the crossed intervention.}
  \label{fig:appendix-layerwise-association}
\end{figure*}

Across 16 OLMoE layers, source-relative hidden-state cosine distance and
Top-$k$ disagreement are positively associated (Pearson $r=0.730$,
$p=0.0013$; Spearman $\rho=0.653$, $p=0.0061$;
Figure~\ref{fig:appendix-layerwise-association}). Both vary with depth;
this association does not replace the crossed intervention.

\subsection{Structural predictors and multiplicity}
\label{app:routing-metric-family}

The eight predictors are set distance, selected probability-mass change,
$\ell_1$ and JS distances between selected weights on their union,
full-distribution JS and symmetric KL, entropy change, and mean absolute
rank change. We compare them with source-policy, source-hidden-state,
source-router, and set-only NLL gains on all events and changed-set events:
256 tests across the four DeepSeekMoE/OLMoE settings.

The frozen primary family comprises 16 changed-set tests: set distance and
full-distribution JS against source-policy and source-hidden-state gains.
Holm correction is applied separately to Spearman and AUROC within each
declared family. None of these primary tests survives correction
(Table~\ref{tab:primary-routing-family}); adjusted primary $p$-values equal
one. The 18 corrected secondary discoveries all involve all-event set
distance versus set-only gain, where an unchanged set mechanically creates
a no-op. They are not treated as independent predictive evidence.

\begin{table*}[t]
\centering
\footnotesize
\setlength{\tabcolsep}{3.2pt}
\caption{Primary AUROC comparisons with 95\% prompt-bootstrap intervals. Route/state denote source-routing/source-hidden-state gains; $0.5$ is chance. None of the 16 comparisons survives Holm correction.}
\label{tab:primary-routing-family}
\begin{tabular}{lcccc}
\toprule
Architecture / Parent & \shortstack{Route / Top-$k$} & \shortstack{Route / JS} & \shortstack{State / Top-$k$} & \shortstack{State / JS} \\
\midrule
DeepSeekMoE / Average & $.490\,[.463,.517]$ & $.510\,[.472,.548]$ & $.502\,[.476,.528]$ & $.509\,[.472,.547]$ \\
DeepSeekMoE / Task Arithmetic & $.491\,[.465,.517]$ & $.510\,[.478,.542]$ & $.503\,[.477,.529]$ & $.511\,[.477,.547]$ \\
OLMoE / Average & $.509\,[.483,.533]$ & $.519\,[.482,.556]$ & $.515\,[.489,.541]$ & $.520\,[.481,.559]$ \\
OLMoE / Task Arithmetic & $.525\,[.502,.548]$ & $.483\,[.450,.515]$ & $.525\,[.502,.549]$ & $.487\,[.454,.519]$ \\
\bottomrule
\end{tabular}
\end{table*}

\begin{figure*}[t]
  \centering
  \includegraphics[width=0.92\textwidth]{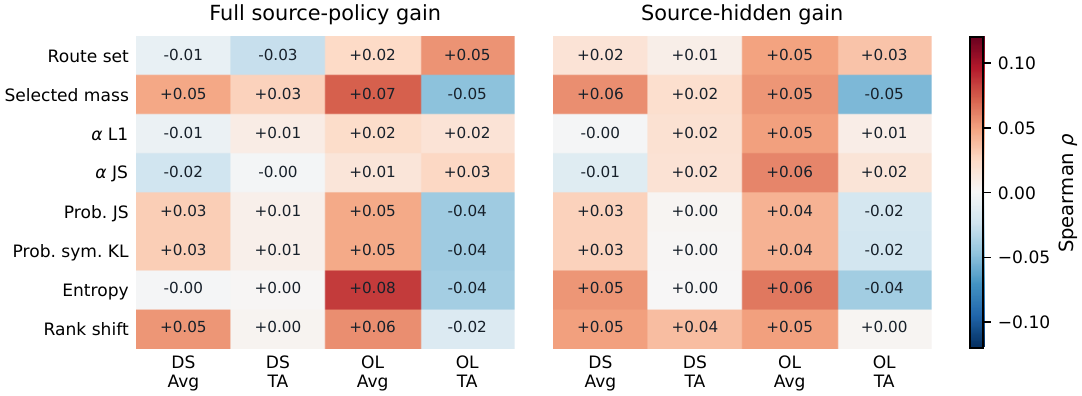}
  \caption{\textbf{Complete changed-set predictor family.} Spearman associations with source-policy gains (left) and source-hidden-state gains (right). Primary predictors are set distance and full-distribution JS.}
  \label{fig:appendix-routing-family}
\end{figure*}

Figure~\ref{fig:appendix-routing-family} displays the complete changed-set
metric family. A separately specified pair/boundary predictor has AUROC
$0.502$, $0.510$, $0.499$, and $0.497$ in the same setting order.

\subsection{Local and task-level source-route replay}
\label{app:task-route-controls}

The token-local experiment replaces one changed route at one layer,
keeping that event's merged input and expert parameters fixed while
subsequent computation responds. Positive $\Delta\mathrm{NLL}$ means
native NLL minus replay NLL. All four intervals in
Table~\ref{tab:token-local-source-route} are negative.

\begin{table}[t]
\centering
\small
\caption{Token-local source-route replay. Positive native-minus-replay NLL favors replay. Intervals resample prompts.}
\label{tab:token-local-source-route}
\begin{tabular}{lcc}
\toprule
Architecture / Parent & $\Delta\mathrm{NLL}$ [95\% CI] & Events / prompts \\
\midrule
DeepSeekMoE / Average & $-0.0339\,[-0.0675,-0.0062]$ & 80 / 71 \\
DeepSeekMoE / Task Arithmetic & $-0.0326\,[-0.0682,-0.0038]$ & 80 / 74 \\
OLMoE / Average & $-0.1623\,[-0.4078,-0.0026]$ & 80 / 66 \\
OLMoE / Task Arithmetic & $-0.1076\,[-0.1947,-0.0400]$ & 80 / 68 \\
\bottomrule
\end{tabular}
\end{table}

For task evaluation, every answer choice is teacher-forced. The source
schedule replaces routing at all non-padding question--choice tokens and
sparse layers, but only answer tokens enter the length-normalized score
in Eq.~\ref{eq:task-grounded-choice-margin}. Science uses ARC-Challenge;
mathematics and computer science each comprise four MMLU subjects with
64 items per subject. Outcome-independent SHA-256 ranking fixes the items,
and each domain uses its matching specialist. All 12 margin and accuracy
intervals in Table~\ref{tab:task-grounded-source-route} include zero.

\begin{table*}[t]
\centering
\footnotesize
\setlength{\tabcolsep}{3pt}
\caption{Source-route task effects on 256 items per row. Margin and accuracy differences favor replay when positive; accuracy is a fraction. Intervals use paired item bootstrap. DS/OL denote DeepSeekMoE/OLMoE.}
\label{tab:task-grounded-source-route}
\begin{tabular}{llcc}
\toprule
Model / Parent & Domain & Margin $\Delta$ [95\% CI] & Accuracy $\Delta$ [95\% CI] \\
\midrule
DS / Average & Science & $-0.0052\,[-0.0182,+0.0079]$ & $+0.0000\,[-0.0156,+0.0156]$ \\
& Mathematics & $-0.0122\,[-0.0259,+0.0020]$ & $-0.0234\,[-0.0547,+0.0078]$ \\
& Computer science & $-0.0005\,[-0.0146,+0.0132]$ & $+0.0234\,[-0.0117,+0.0586]$ \\
DS / Task Arithmetic & Science & $+0.0021\,[-0.0118,+0.0170]$ & $+0.0078\,[-0.0078,+0.0234]$ \\
& Mathematics & $-0.0107\,[-0.0259,+0.0044]$ & $-0.0273\,[-0.0664,+0.0117]$ \\
& Computer science & $-0.0005\,[-0.0161,+0.0146]$ & $+0.0312\,[-0.0039,+0.0664]$ \\
OL / Average & Science & $+0.0025\,[-0.0261,+0.0312]$ & $+0.0117\,[-0.0156,+0.0391]$ \\
& Mathematics & $+0.0046\,[-0.0342,+0.0442]$ & $-0.0117\,[-0.0547,+0.0273]$ \\
& Computer science & $+0.0200\,[-0.0266,+0.0652]$ & $+0.0195\,[-0.0078,+0.0469]$ \\
OL / Task Arithmetic & Science & $-0.0062\,[-0.0418,+0.0285]$ & $+0.0117\,[-0.0117,+0.0352]$ \\
& Mathematics & $+0.0525\,[-0.0063,+0.1130]$ & $+0.0000\,[-0.0430,+0.0430]$ \\
& Computer science & $+0.0034\,[-0.0605,+0.0664]$ & $-0.0117\,[-0.0469,+0.0195]$ \\
\bottomrule
\end{tabular}
\end{table*}

\subsection{Qwen3-MoE measurements}
\label{app:qwen-route-diagnosis}

Qwen3-30B-A3B-Base is evaluated under Average and TA with coefficient
$0.5625$. Its 256 trajectories per parent retain an earlier 64-prompt
sample and add 192 greedy continuations capped at 96 new tokens.
Additional selection rules were fixed before collecting the added outcomes,
but this extension followed earlier diagnostics; it does not change the
original 16-test primary family.

Eight hash-selected events per prompt are drawn from layers 43--47 without
duplicating short trajectories. Average yields 2,045 events and TA 2,048;
the changed-set subsets contain 725 events from 235 prompts and 858 from
240 prompts. Table~\ref{tab:qwen-routing-diagnosis} retains all three
available metrics. Its 4,000-draw prompt-bootstrap intervals are pointwise:
all correlation intervals include zero and all AUROC intervals include
$0.5$.

\begin{table*}[t]
\centering
\footnotesize
\setlength{\tabcolsep}{4pt}
\caption{Qwen3 structural diagnostics on changed-set events. Intervals are pointwise 95\% prompt-bootstrap intervals. Selected-weight JS aligns weights on the union of selected experts.}
\label{tab:qwen-routing-diagnosis}
\begin{tabular}{llcc}
\toprule
Parent & Distance & Spearman $\rho$ [95\% CI] & AUROC [95\% CI] \\
\midrule
Average & Full-routing JS & $-.036\,[-.112,.038]$ & $.474\,[.431,.518]$ \\
& Top-$k$ set & $.015\,[-.061,.092]$ & $.506\,[.473,.540]$ \\
& Selected-weight JS & $-.022\,[-.099,.058]$ & $.482\,[.437,.530]$ \\
Task Arithmetic & Full-routing JS & $.028\,[-.045,.103]$ & $.514\,[.475,.553]$ \\
& Top-$k$ set & $-.009\,[-.081,.058]$ & $.500\,[.469,.530]$ \\
& Selected-weight JS & $.037\,[-.039,.108]$ & $.522\,[.483,.561]$ \\
\bottomrule
\end{tabular}
\vspace{-2em}
\end{table*}

The all-layer attribution uses every continuation-predicting position and
all 48 sparse layers, selecting eight of 128 experts. Average and TA
contribute 1,081,728 and 1,105,824 records, respectively. Attribution ratios
are recomputed in 10,000 prompt resamples, rather than averaging prompt
percentages (Table~\ref{tab:qwen-routing-origin}).

\begin{table*}[b]
\centering
\small
\setlength{\tabcolsep}{5pt}
\caption{Qwen3 routing attribution (\%; 95\% prompt-bootstrap intervals). The first row includes all events; remaining rows condition on a source--merged set change.}
\label{tab:qwen-routing-origin}
\begin{tabular}{lcc}
\toprule
Quantity & Average & Task Arithmetic \\
\midrule
Overall flip rate & $26.46\,[25.01,27.96]$ & $33.20\,[31.48,34.94]$ \\
Representation-induced & $96.87\,[96.75,96.98]$ & $96.90\,[96.80,97.01]$ \\
Router-parameter-induced & $0.034\,[0.027,0.041]$ & $0.040\,[0.033,0.047]$ \\
Interaction-required & $0.929\,[0.873,0.989]$ & $0.774\,[0.728,0.823]$ \\
Redundantly sufficient & $2.171\,[2.086,2.259]$ & $2.282\,[2.207,2.357]$ \\
\bottomrule
\end{tabular}
\end{table*}

Task replay uses the same 64 ARC-Challenge items and 256 answer choices
for both parents, with interventions at all non-padding tokens and all
sparse layers. Each parent loses one correct answer and gains none;
10,000 paired item resamples give the intervals below. No mathematics or
computer-science replay is included in this Qwen3 sample.

\begin{table*}[b]
\centering
\small
\caption{Qwen3 source-route replay on 64 ARC-Challenge items. Positive changes favor replay; accuracy changes are fractions. Intervals use paired item bootstrap.}
\label{tab:qwen-task-restoration}
\begin{tabular}{lcc}
\toprule
Parent & Margin $\Delta$ [95\% CI] & Accuracy $\Delta$ [95\% CI] \\
\midrule
Average & $+.00880\,[-.01196,+.02822]$ & $-.015625\,[-.046875,0]$ \\
Task Arithmetic & $+.00180\,[-.01685,+.01968]$ & $-.015625\,[-.046875,0]$ \\
\bottomrule
\end{tabular}
\end{table*}

\subsection{Local reference quality}
\label{app:local-reference-quality}

Each event has 7--12 unique candidate routes: native, source, the two crossed
states in Eq.~\ref{eq:four-state-routing}, a frozen linear-calibration
route, and deterministic one-expert swaps. Candidates use identical merged
inputs and experts. ``Source wins'' is a strict NLL comparison; ``Merged
best'' admits ties within $10^{-6}$. Table~\ref{tab:local-reference-quality}
shows limited native optimality without a reliable source advantage. This
is a bounded candidate comparison, not an exhaustive route search.

\begin{table*}[t]
\centering
\footnotesize
\setlength{\tabcolsep}{3pt}
\caption{NLL reference quality within the fixed local candidate set. Source wins are strict; merged-best counts admit ties within $10^{-6}$. Intervals resample prompts.}
\label{tab:local-reference-quality}
\begin{tabular}{lcccc}
\toprule
Model / Parent & Source wins [95\% CI] & Ties & Merged best [95\% CI] & Candidates, mean [range] \\
\midrule
DS / Average & $49.6\,[43.4,55.9]\%$ & $3.1\%$ & $10.9\,[7.0,14.8]\%$ & $9.34\,[7,12]$ \\
DS / Task Arithmetic & $48.8\,[43.0,55.1]\%$ & $0.4\%$ & $12.9\,[9.0,17.2]\%$ & $9.69\,[7,12]$ \\
OL / Average & $50.0\,[43.8,56.2]\%$ & $0.8\%$ & $13.3\,[9.4,17.6]\%$ & $9.31\,[7,12]$ \\
OL / Task Arithmetic & $45.7\,[39.8,52.0]\%$ & $2.0\%$ & $9.8\,[6.2,13.7]\%$ & $9.55\,[7,12]$ \\
\bottomrule
\end{tabular}
\end{table*}

\subsection{Mixture outputs and matched controls}
\label{app:changed-route-redundancy}

One event per prompt is hash-selected from the final-five-layer pool before
classifying route changes. Within each setting, expert-pair, observed-mixture,
and one-swap measurements use the same changed events
(Table~\ref{tab:changed-route-redundancy}). Expert-pair similarity is the
maximum entering--leaving output cosine. Mixtures use identical merged
inputs and expert parameters; DeepSeekMoE's fixed shared-expert branch is
excluded. The 727 event-wise mixture-minus-expert gaps are all positive,
with range $[0.078,1.417]$. Gaps can exceed one because an expert-output
cosine can be negative; no tail events are discarded.

\begin{table*}[t]
\centering
\footnotesize
\setlength{\tabcolsep}{3pt}
\caption{Directional mixture comparisons on matched changed-route events. Values are mean cosines with 95\% prompt-bootstrap intervals. DS/OL/QW denote DeepSeekMoE/OLMoE/Qwen3-MoE.}
\label{tab:changed-route-redundancy}
\begin{tabular}{lcccc}
\toprule
Model / Parent & Events & Best expert pair & Source-route mixture & One-swap mixture \\
\midrule
DS / Average & 115 & $0.050\,[0.035,0.067]$ & $0.913\,[0.887,0.936]$ & $0.951\,[0.936,0.964]$ \\
DS / Task Arithmetic & 149 & $0.043\,[0.031,0.056]$ & $0.896\,[0.867,0.923]$ & $0.962\,[0.950,0.972]$ \\
OL / Average & 112 & $0.074\,[0.050,0.101]$ & $0.976\,[0.969,0.982]$ & $0.990\,[0.988,0.992]$ \\
OL / Task Arithmetic & 142 & $0.084\,[0.062,0.108]$ & $0.965\,[0.956,0.972]$ & $0.987\,[0.984,0.989]$ \\
QW / Average & 91 & $0.041\,[0.019,0.065]$ & $0.953\,[0.936,0.967]$ & $0.980\,[0.974,0.984]$ \\
QW / Task Arithmetic & 118 & $0.096\,[0.066,0.132]$ & $0.957\,[0.940,0.970]$ & $0.975\,[0.964,0.983]$ \\
\bottomrule
\end{tabular}
\end{table*}

Unchanged-route references are the complementary selected events, not
additional interventions on the changed-event sample. Their reported
mixture cosines are retained in Table~\ref{tab:unchanged-route-reference}.

\begin{table}[t]
\centering
\small
\caption{Mixture cosine on unchanged-route reference events. These events
are complementary to, not matched replicates of, the changed-route sample.
Values are means with 95\% prompt-bootstrap intervals.}
\label{tab:unchanged-route-reference}
\begin{tabular}{@{}lcc@{}}
\toprule
Architecture & Average & Task Arithmetic \\
\midrule
DeepSeekMoE & $0.997\,[0.994,0.999]$ & $0.998\,[0.997,0.999]$ \\
OLMoE & $0.997\,[0.996,0.998]$ & $0.996\,[0.995,0.997]$ \\
Qwen3-MoE & $1.000\,[0.999,1.000]$ & $1.000\,[0.999,1.000]$ \\
\bottomrule
\end{tabular}
\end{table}

\paragraph{One-expert swaps.}
The lowest-logit selected expert is replaced from the parent/source
Top-$(k+4)$ union. Parent/source ranks order at most seven alternatives,
independently of their outputs. Parent probabilities on the candidate set
are scaled to native routed mass and stored in BF16. We average within
events before averaging prompts. Qwen3 excludes native/crossed sets;
the DeepSeekMoE/OLMoE list additionally excludes its LC candidate. This
reference-family difference does not enter the matched-random contrast.

\paragraph{Matched random routes.}
Each event has 32 alternatives preserving cardinality, the shared subset
with the merged route, the exact BF16 source-weight multiset, routed mass,
and concentration. Hashes determine expert alternatives and weight-slot
assignments without inspecting outputs. The paired contrast subtracts
mean random-route cosine from observed cosine within each event, followed
by 10,000 prompt resamples. DeepSeekMoE/OLMoE random controls use a separate
explicit-dispatch run; their absolute observed means must not be replaced
by those in Table~\ref{tab:changed-route-redundancy}. Qwen3 uses one run
for both comparisons. Table~\ref{tab:matched-random-route} preserves these
within-run quantities. Cosine measures direction, not output magnitude
or task recovery.

\begin{table*}[t]
\centering
\small
\caption{Observed source routing versus 32 matched random alternatives per event. Intervals apply to the paired observed-minus-random cosine, not separately to the absolute means.}
\label{tab:matched-random-route}
\begin{tabular}{lccc}
\toprule
Architecture / Parent & Observed source & Matched random & Difference [95\% CI] \\
\midrule
DeepSeekMoE / Average & 0.902 & 0.769 & $+0.133\,[+0.109,+0.158]$ \\
DeepSeekMoE / Task Arithmetic & 0.895 & 0.748 & $+0.147\,[+0.127,+0.168]$ \\
OLMoE / Average & 0.976 & 0.855 & $+0.122\,[+0.102,+0.143]$ \\
OLMoE / Task Arithmetic & 0.966 & 0.854 & $+0.112\,[+0.095,+0.129]$ \\
Qwen3-MoE / Average & 0.953 & 0.840 & $+0.113\,[+0.099,+0.128]$ \\
Qwen3-MoE / Task Arithmetic & 0.957 & 0.841 & $+0.116\,[+0.102,+0.131]$ \\
\bottomrule
\end{tabular}
\end{table*}

\subsection{Native routing and figure conventions}
\label{app:native-routing-controls}

OLMoE uses full softmax followed by Top-8, without additional renormalization
unless required by its configuration. DeepSeekMoE uses Top-6 and preserves
the shared branch through routed-output delta replacement. Qwen3 uses
FP32 softmax, sorted Top-8, native renormalization, and BF16 dispatch weights.
The four DeepSeekMoE/OLMoE parents and eight specialists pass native-route
round trips with zero NLL and output-recomputation error; the 12 corresponding
task cells also reproduce native routing. An OLMoE TA boundary audit records
one tied one-swap discrepancy, also present in source self-replay.

For Qwen3, unsorted replay failed the original tolerance and was replaced
by native sorted dispatch before the expanded diagnostic outcomes were
inspected. Final functional and task replays report zero reconstruction
error under the unchanged $2\times10^{-3}$ tolerance. Mixture comparisons
capture selected experts from their original native dispatch and evaluate
unselected experts at the same saved input, using one output table for all
routes. Raw batched recomputation differs by up to $0.002727$ in relative
norm; captured native outputs reconstruct exactly.

Figure~\ref{fig:mixture-redundancy} classifies Qwen3 events using native GPU
selections. Exact BF16 boundary ties cause 16/17 CPU--GPU discrepancies and
9/7 changed-set reclassifications for Average/TA. In contrast,
Figure~\ref{fig:route-diagnosis} retains its historical CPU route labels.
Earlier and added events retain their respective source-route caches.
These conventions, as well as the sampling frames, must be retained when
comparing counts across figures.

Figure~\ref{fig:route-origin}(a) reuses 64 outcome-independent event identities
across its three replacement rows within each setting; columns are not
matched across architectures. Its color scale is linear on $[0,1]$.
The displayed set distance is $1-|S_{\rm src}\cap S_{\rm cond}|/k$;
Qwen3's functional set-distance analysis uses one minus Jaccard similarity.
Figure~\ref{fig:route-diagnosis} retains all 5,762 finite changed-set
observations without jitter, added epsilon, or clipping. JS uses a log
axis and gains a symmetric-log axis, with linear region $[-0.002,0.002]$.
Ranks and AUROC are computed from raw values, with larger JS predicting
positive gain; plotted transformations do not change the tests.

\subsection{Variation outside merging}
\label{app:merge-specificity}

Source--base set distances exceed source--merged distances, while different
specialist routers on the same source state still disagree
(Table~\ref{tab:merge-specificity}). Routing variation therefore also
occurs outside merging. These controls qualify the setting-specific
interpretation without changing the local test.

\begin{table}[tbp]
\centering
\small
\setlength{\tabcolsep}{5pt}
\caption{Set-distance controls outside merging, with 95\% prompt-bootstrap intervals.}
\label{tab:merge-specificity}
\begin{tabular}{@{}lcc@{}}
\toprule
Contrast & DeepSeekMoE & OLMoE \\
\midrule
Source--Average & $0.149\,[0.138,0.160]$ & $0.114\,[0.106,0.122]$ \\
Source--TA & $0.209\,[0.195,0.224]$ & $0.159\,[0.150,0.168]$ \\
Base--source & $0.271\,[0.255,0.287]$ & $0.215\,[0.204,0.226]$ \\
Source--source router only & $0.058\,[0.053,0.063]$ & $0.022\,[0.019,0.024]$ \\
\bottomrule
\end{tabular}
\end{table}

\FloatBarrier

\section{SRR Algorithm and Fitting Objective}
\label{app:srr-algorithm}

Unlike model fusion \citep{gu2026infifpo, wang2026infigfusion}, 
SRR changes only selected rows of the merged parent's router; experts and all other parameters remain fixed.
SRR is a source-informed \emph{candidate construction}, not a test that
source routing is beneficial. It changes only selected rows of the merged
parent's router; experts and all other parameters remain fixed. The same
parent-generated continuations are scored by the base, each domain specialist,
and the parent, each using its own hidden states (Section~\ref{sec:srr}).

\noindent\textbf{SRR construction (pseudocode).}
\par\smallskip
\begingroup
\small
\setlength{\tabcolsep}{4pt}
\begin{tabularx}{\linewidth}{@{}rX@{}}
\toprule
\textbf{Input} & Base $B$, specialists $\{S_d\}$, merged parent $M$,
domain-labeled prompts, and fixed SRR settings. \\
\midrule
1 & Generate continuations with $M$ for fixed construction-train and
validation prompt splits. Cache aligned token scores and router traces
from $B$, $S_d$, and $M$. \\
2 & In each of the final five sparse layers, compute the weighted
source--base prompt profiles (Eq.~\ref{eq:srr-main-profile}). Require sign
consistency $\geq0.75$ and activity $\geq0.002$ in both splits; select two
disjoint expert pairs per domain. \\
3 & For each selected pair $j$, form the clipped source--parent target
$\widetilde r_{tj}$, reliability weight $\rho_{tj}$, and parent input $h_t^M$
on construction-train tokens. Require positive-weight support. \\
4 & Solve the pairwise weighted ridge objective below in FP64; stop without
a candidate if a pair fails to converge. \\
5 & Apply all equal-and-opposite pair updates jointly to the FP32 parent
router, then verify the changed-layer scope after BF16 conversion. \\
\textbf{Output} & SRR candidate checkpoint and fit audit; no benchmark
score enters selection or fitting. \\
\bottomrule
\end{tabularx}
\endgroup

\paragraph{Complete fitting objective.}
For pair $j=(a_j,b_j)$ assigned to domain $d_j$, let
$S_t=S_{d_t}$, $m_{tj}^{A}=z_{t,a_j}^{A}-z_{t,b_j}^{A}$, and
$\delta_t=n_t^M-n_t^{S_t}$. The target and weight are
\begin{equation}
\begin{aligned}
r_{tj}&=m_{tj}^{S_t}-m_{tj}^{M},\\
\widetilde r_{tj}&=\operatorname{sgn}(r_{tj})
\min\{|r_{tj}|,|m_{tj}^{S_t}-m_{tj}^{B}|\},\\
\rho_{tj}&=\mathbf1[d_t=d_j]\mathbf1[\delta_t>0]
\sigma(\delta_t/\tau).
\end{aligned}
\label{eq:srr-app-target}
\end{equation}
For $N$ pooled construction-train response tokens, including those with
zero weight, the independent fit for each pair is
\begin{equation}
\begin{aligned}
v_j^\star&=\arg\min_v\;\frac12\sum_{t=1}^{N}\rho_{tj}
\big((h_t^M)^\top v-\widetilde r_{tj}\big)^2
+\frac{\lambda_j}{2}\|v\|_2^2,\\
\lambda_j&=\gamma\max\!\left\{\frac1N\sum_{t=1}^{N}
\rho_{tj}\|h_t^M\|_2^2,10^{-12}\right\}.
\end{aligned}
\label{eq:srr-app-objective}
\end{equation}
Nonmatching domains receive \emph{zero weight}, not a zero-change target.
The fitted vector changes rows $a_j$ and $b_j$ by
$+\eta v_j^\star/2$ and $-\eta v_j^\star/2$, respectively
(Eq.~\ref{eq:srr-main-update}).

\paragraph{Fit and execution.}
We use $\tau=0.5$, $\gamma=10^{-3}$, and $\eta=0.0625$ throughout.
Diagonal-preconditioned conjugate gradients solves each ridge system in
FP64, checking the true relative residual against $10^{-6}$ within at
most 4096 iterations. All fits use unchanged parent traces and precede
joint writeback; there is no gradient-training epoch or trajectory refresh.
The validation split screens pairs and audits the fixed-input logit change,
but does not tune the update scale. Pair scarcity, absent positive support,
solver failure, or an incorrect BF16 changed-layer scope aborts construction
rather than triggering a benchmark-dependent fallback. The objective
matches selected router-logit gaps, not task loss; successful fitting alone
does not imply downstream recovery (Appendix~\ref{app:srr-counterexample}).
Further selection and solver details appear in
Appendix~\ref{app:srr-implementation}.

\section{Task-Grounded Intervention Protocols}
\label{app:task-grounded-failure}

This appendix reports the scoring convention, controlled corruption, and
natural LC replay used in Section~\ref{sec:task-grounded-failure}.
Source-specialist task replay is reported separately in
Appendices~\ref{app:task-route-controls} and~\ref{app:qwen-route-diagnosis}.
Each effect is paired within its own model, item set, and routing reference.

\subsection{Task scoring and prediction transitions}
\label{app:task-scoring}

For choice $c$ with answer tokens $y_c$, define
\begin{equation}
s_c=-\frac{1}{|y_c|}\sum_t\operatorname{NLL}(y_{c,t}),
\qquad m=s_{c^*}-\max_{c\ne c^*}s_c.
\label{eq:task-grounded-choice-margin}
\end{equation}
The prediction maximizes $s_c$; only answer tokens contribute to scoring.
Task labels evaluate a specified routing intervention, not a separately
chosen reference for each correct answer. Margin is a continuous score
readout, distinct from accuracy. For $N$ paired items,
\begin{equation}
\Delta\operatorname{Acc}
=\frac{n_{\mathrm{fix}}-n_{\mathrm{break}}}{N},
\label{eq:task-grounded-transitions}
\end{equation}
where the counts denote wrong-to-correct and correct-to-wrong transitions.
Multiplying by 100 converts accuracy differences to percentage points.
Intervals use 10,000 paired item-bootstrap draws.

\subsection{Full-layer and graded corruption controls}
\label{app:corruption-controls}

The original control permutes OLMoE expert logits at layers 0--15 and
replays the same parent's known clean schedule. The fixed seeds are
2026091701 (Average) and 2026091702 (TA), with 256 items per parent.
Both comparisons increase accuracy from $22.65625\%$ to $50.78125\%$,
corresponding to 72 net corrections, but not necessarily identical
prediction transitions. Table~\ref{tab:task-grounded-details} preserves
accuracy and margin effects. The original injected-corruption comparisons
correct 94 and 90 answers and break 22 and 18 under Average and TA,
respectively; the separate natural LC comparisons correct/break 0/1 and 1/0.

\begin{table}[t]
\centering
\small
\setlength{\tabcolsep}{5pt}
\renewcommand{\arraystretch}{1.10}
\caption{Original OLMoE replay controls. Accuracy changes are percentage points; margins retain score units. Brackets are 95\% paired item-bootstrap intervals. The injected and LC comparisons are distinct experiments.}
\label{tab:task-grounded-details}
\label{tab:task-grounded-failure}
\begin{tabular}{@{}lcc@{}}
\toprule
Parent & Accuracy change [interval] & Margin change [interval] \\
\midrule
\multicolumn{3}{@{}l}{\textit{Injected corruption $\rightarrow$ clean-native routing replay}} \\
Average & $+28.125\;[+20.313,\,+35.547]$ & $+0.7677\;[+0.5411,\,+0.9934]$ \\
Task Arithmetic & $+28.125\;[+21.094,\,+35.156]$ & $+0.5588\;[+0.3675,\,+0.7549]$ \\
\addlinespace[4pt]
\multicolumn{3}{@{}l}{\textit{Native merged routing $\rightarrow$ frozen LC routing replay}} \\
Average & $-0.391\;[-1.172,\,0.000]$ & $-0.0003\;[-0.0037,\,+0.0032]$ \\
Task Arithmetic & $+0.391\;[0.000,\,+1.172]$ & $-0.0026\;[-0.0066,\,+0.0008]$ \\
\bottomrule
\end{tabular}
\end{table}

The graded experiment uses the same archived parents and 256 matched ARC
items, but a separate randomization. Seeds 17017 and 29029 fix nested layer
orders; the first $k=1,4,16$ layers receive a logit permutation.
Table~\ref{tab:graded-router-corruption} retains every condition,
pointwise paired interval, and the Holm decision across twelve tests.
All eight four- and sixteen-layer recoveries pass the specified correction;
none of the one-layer tests does. Seeds are fixed corruption realizations,
not independent trained models. Replay ranges from $50.39\%$ to $51.56\%$
around the $50.78\%$ clean reference and does not establish exact
reconstruction. These controls concern the imposed losses, not learned SRR
repair or a universal sensitivity threshold.

\begin{table*}[t]
\centering
\footnotesize
\setlength{\tabcolsep}{3pt}
\renewcommand{\arraystretch}{1.04}
\caption{Graded OLMoE corruption on 256 items per parent. Accuracy is in \%; recovery and its paired 95\% interval are in pp. Clean accuracy is 50.78\%. Holm decisions cover all twelve parent--seed--scope comparisons.}
\label{tab:graded-router-corruption}
\begin{tabular}{@{}llrrrlc@{}}
\toprule
Parent & Permutation seed & $k$ & Corrupted & Replay & Recovery [95\% CI] & Holm $p<.05$ \\
\midrule
Average & 17017 & 1  & 53.12 & 50.78 & $-2.34\;[-6.25,+1.56]$ & No \\
        &       & 4  & 38.28 & 50.78 & $+12.50\;[+7.03,+18.36]$ & Yes \\
        &       & 16 & 17.97 & 50.78 & $+32.81\;[+25.39,+40.23]$ & Yes \\
        & 29029 & 1  & 49.22 & 50.39 & $+1.17\;[-3.52,+5.86]$ & No \\
        &       & 4  & 37.89 & 50.78 & $+12.89\;[+6.64,+18.76]$ & Yes \\
        &       & 16 & 23.44 & 50.78 & $+27.34\;[+19.92,+34.77]$ & Yes \\
\addlinespace[2pt]
TA      & 17017 & 1  & 52.73 & 50.78 & $-1.95\;[-5.86,+1.95]$ & No \\
        &       & 4  & 36.33 & 51.17 & $+14.84\;[+9.38,+20.70]$ & Yes \\
        &       & 16 & 19.92 & 51.56 & $+31.64\;[+23.83,+39.45]$ & Yes \\
        & 29029 & 1  & 48.83 & 51.17 & $+2.34\;[-1.95,+6.64]$ & No \\
        &       & 4  & 36.72 & 51.17 & $+14.45\;[+8.59,+20.31]$ & Yes \\
        &       & 16 & 24.61 & 51.17 & $+26.56\;[+18.75,+33.98]$ & Yes \\
\bottomrule
\end{tabular}
\end{table*}

\begin{figure}[t]
\centering
\includegraphics[width=\textwidth]{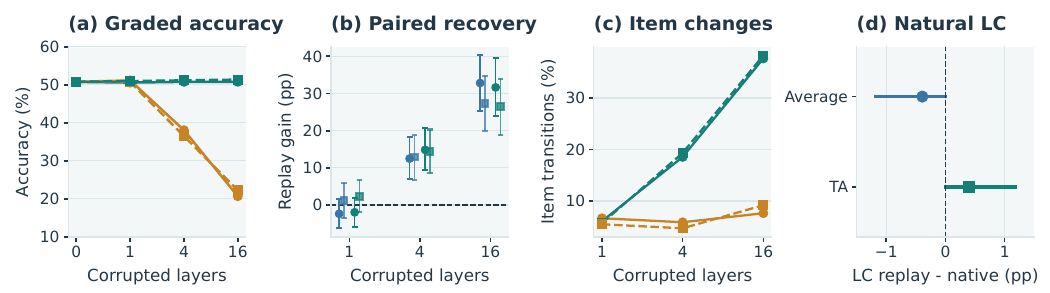}
\caption{\textbf{Graded routing interventions.} (a) Clean, corrupted, and replay accuracy, averaging two fixed permutations per parent. (b) All twelve paired recovery effects with 95\% item-bootstrap intervals. (c) Corrections and regressions, averaged within parent and scope. (d) Independent natural LC effects. Seeds are not model replicates.}
\label{fig:full-task-diagnostics}
\end{figure}

\subsection{Natural LC replay and numerical validity}
\label{app:lc-replay}

A separate OLMoE cohort replays schedules from frozen LC-MERGE-RC router
checkpoints on 256 ARC items per parent. All non-routing parameters remain
those of the native parent. The archive name \emph{LC} distinguishes this
fixed-sequence replay from end-to-end SRR evaluation. Accuracy changes by
$-0.390625$ pp for Average and $+0.390625$ pp for TA: one changed-correctness
item per parent in opposite directions. Both accuracy intervals include
zero; both mean margin changes are negative with intervals crossing zero.
The natural and injected cohorts are not pooled.


The original positive-control bootstrap estimates and transition counts were
reproduced from the archived item records and recorded seeds, without model
inference. The OLMoE LC identity replay reports zero native-recomputation
relative error under tolerance $2\times10^{-3}$.
This DeepSeekMoE check concerns only the separate LC
schedule-replay experiment, not the end-to-end SRR evaluation in
Table~\ref{tab:main-results}.

\FloatBarrier

\section{SRR Implementation and Proxy Checks}
\label{app:srr-implementation}

Section~\ref{sec:srr} defines the profile, target, regression, and writeback.
Here we specify the remaining selection and numerical details, then retain
two checks distinguishing proxy fit from task recovery. The fixed values
are collected in Table~\ref{tab:srr-configuration}; none is selected using
benchmark outcomes.

\begin{table}[tbp]
\centering
\footnotesize
\setlength{\tabcolsep}{6pt}
\caption{Fixed SRR settings. Pair quotas apply to each domain at each eligible layer.}
\label{tab:srr-configuration}
\begin{tabular}{@{}lrlr@{}}
\toprule
Quantity & Value & Quantity & Value \\
\midrule
Likelihood temperature $\tau$ & $0.5$ & Relative ridge $\gamma$ & $10^{-3}$ \\
Sign consistency $\kappa$ & $0.75$ & Update scale $\eta$ & $0.0625$ \\
Minimum activity $a_{\min}$ & $0.002$ & Profile/activity floor & $10^{-8}$ \\
Pairs per domain/layer & $2$ & Ridge energy floor & $10^{-12}$ \\
Eligible router layers & Final $5$ & PCG iteration limit & $4096$ \\
True-residual tolerance & $10^{-6}$ & & \\
\bottomrule
\end{tabular}
\end{table}

\subsection{Deterministic expert-pair selection}
\label{sec:srr-pairs}

The implementation forms the profile from centered log-softmax differences.
Centering removes the common offset, giving
$\operatorname{center}(\log\operatorname{softmax}(z^S)
-\log\operatorname{softmax}(z^B))=\operatorname{center}(z^S-z^B)$
in exact arithmetic. Unlike fitting weights, profile weights $\omega_t$
are not thresholded at positive advantage. Expert activity is
\begin{equation}
a_{ie}=
\frac{\sum_{t\in i}\omega_t
\max\{\operatorname{softmax}(z_t^S)_e,
       \operatorname{softmax}(z_t^B)_e\}}
{\max\{\sum_{t\in i}\omega_t,10^{-8}\}}.
\label{eq:srr-activity}
\end{equation}
The same denominator floor applies to the prompt profile in
Eq.~\ref{eq:srr-main-profile}.

For each domain and split $s\in\{\mathrm{tr},\mathrm{val}\}$, let
$\mu_e^s$ and $A_e^s$ be prompt means of the profile and activity, and
$f_{e,+}^s,f_{e,-}^s$ the fractions with positive and negative profile.
An eligible positive expert satisfies $\mu_e^s>0$,
$f_{e,+}^s\ge\kappa$, and $A_e^s\ge a_{\min}$ in both splits;
negative eligibility is symmetric. Pairs $(a,b)$ are ranked by
\begin{equation}
q_{ab}=\min_s(\mu_a^s-\mu_b^s)
\min_s\{f_{a,+}^s,f_{b,-}^s\}
\sqrt{\min_s\{A_a^s,A_b^s\}}.
\label{eq:srr-pair-ranking}
\end{equation}
Expert indices break ties. In fixed domain order, selection takes the
highest-ranked unused pair, without reusing endpoints within a layer.
Construction stops if a domain cannot meet its quota or a selected pair
has no positive source-over-parent training support; there is no
benchmark-dependent fallback.

\subsection{Numerical solve and writeback}
\label{sec:srr-corrections}

Let $H$ stack training inputs and $D_j=\operatorname{diag}(\rho_{\cdot j})$.
The regression in Eq.~\ref{eq:srr-main-fit} solves
\begin{equation}
(H^\top D_jH+\lambda_j I)v_j=H^\top D_j\widetilde r_j.
\label{eq:srr-normal-equations}
\end{equation}
Its data term is a weighted sum. The $N$ in the ridge scale counts all
pooled training response tokens, including zero-weight tokens for that pair.
Nonmatching-domain tokens have zero weight, not a positively weighted
zero-change preservation target.

Batched FP64 preconditioned conjugate gradients maintains a separate
convergence state per pair. Matrix products use
$H^\top[\rho_{\cdot j}\odot(Hv)]+\lambda_jv$ and the diagonal
$\sum_t\rho_{tj}(h_t^M\odot h_t^M)+\lambda_j\mathbf1$ as preconditioner.
The true system residual is recomputed, and every selected pair must
converge before writeback. All fits use unchanged parent traces; updates
are then applied jointly. FP32 router weights are saved and changed-layer
scope is checked after BF16 conversion. The covariance-based logit-change
budget is diagnostic, not an enforced projection. Sparse row updates do
not guarantee unchanged probabilities or expert choices elsewhere.

\subsection{Held-out weighting ablation}
\label{app:srr-weighting-ablation}

A separate DeepSeekMoE/OLMoE Average--TA experiment varies the residual-fitting
weights. Table~\ref{tab:task-grounded-weighting} reports all-held-out MSE,
not task performance. All-uniform has the lowest reported MSE in all four
settings; weighted-positive beats shuffled-positive in one. These summaries
lack paired prompt uncertainty and complete support-normalization details,
so they do not establish the task-level superiority of any weighting rule.
This experiment is separate from the local intervention sample in
Appendix~\ref{app:source-interventions}.

\begin{table}[htbp]
\centering
\footnotesize
\setlength{\tabcolsep}{3pt}
\renewcommand{\arraystretch}{1.08}
\caption{Held-out residual-fitting MSE in a separate weighting ablation; lower is better. These point estimates are not task scores and have no paired uncertainty. DS/OL denote DeepSeekMoE/OLMoE.}
\label{tab:task-grounded-weighting}
\begin{tabular}{@{}lrrrr@{}}
\toprule
Cell & Weighted positive & Uniform positive & All uniform & Shuffled positive \\
\midrule
DS / Average & .01774 & .01682 & .01624 & .01761 \\
DS / TA & .05171 & .05065 & .04973 & .05406 \\
OL / Average & .06645 & .05756 & .05471 & .06211 \\
OL / TA & .11473 & .10365 & .09864 & .10738 \\
\bottomrule
\end{tabular}
\end{table}

\subsection{A counterexample to task improvement from proxy fitting}
\label{app:srr-counterexample}

Consider a hypothetical normalized Top-1 layer with $h=1$,
$z^M=z^B=(0,\epsilon)$ and $z^S=(2\epsilon,0)$, where $\epsilon>0$.
Parent experts $b$ and $a$ assign the correct token probabilities $0.8$
and $0.1$; source expert $a$ assigns $0.9$, and the base may assign $0.6$.
The source therefore has positive likelihood advantage, while the selected
pair residual and its clipping bound both equal $3\epsilon$.
With one fitting token, $\gamma=1$ and an inactive numerical floor,
$\lambda=\rho>0$ and $v^\star=3\epsilon/2$. At $\eta=1$, the new parent
logits are $(3\epsilon/4,\epsilon/4)$: the fitting error falls to one
quarter, but routing switches to the worse merged expert and NLL rises by
$\log 8$. This construction is not an experiment or a claim about the
frozen hyperparameters. It shows why the proxy alone supplies no general
task-recovery guarantee when source and merged expert functions differ.

\FloatBarrier

\section{Local Utility of Source-Informed Corrections}
\label{app:source-interventions}

\subsection{Independent sample and paired contrasts}
\label{app:local-sample}

Related on-policy distillation work cautions that token-level disagreement need not be learnable \citep{wang2026not}; 
here we test the distinct question of whether source-likelihood advantage identifies beneficial router corrections.
The local diagnostic covers the four OLMoE/Qwen3-MoE Average--TA settings,
using 256 independent prompts per candidate and one hash-selected
token--layer--pair event per prompt. Positive source advantage occurs in
101, 122, 117, and 131 events, respectively: 471 of 1,024 events.
Construction validation is used for pair selection, not as this independent
diagnostic sample.

Each event compares native routing with equal-magnitude positive and opposite
logit perturbations in the raw source-informed and fitted SRR directions.
The $U$ and $D$ contrasts follow Eq.~\ref{eq:srr-local-contrasts}; $D>0$
does not by itself establish improvement over native. Intervals use 10,000
prompt-bootstrap draws with the fixed analysis seed. The tables retain
positive-support, non-positive-support, and all-event results separately.
No local NLL result is converted into task accuracy.

\begin{table*}[tbp]
\centering
\footnotesize
\setlength{\tabcolsep}{4pt}
\renewcommand{\arraystretch}{1.13}
\caption{\textbf{Local intervention gains by support group.}
Entries are estimates and 95\% prompt-bootstrap intervals in
$10^{-3}$ nats/token. $U$ compares the proposed direction with native
routing; $D$ compares it with the opposite direction. Support groups are
reported separately, not pooled as independent samples.}
\label{tab:p1-supports-compact}
\begingroup
\newcommand{\pione}[2]{\shortstack{$#1$\\[-1pt]$[#2]$}}
\arrayrulecolor{RTRule}
\begin{tabular}{@{}lrcccc@{}}
\toprule
\rowcolor{RTHeader}
Setting & $n$ & Source $U$ & Source $D$ & Fitted $U$ & Fitted $D$ \\
\midrule
\rowcolor{RTBlue}
\multicolumn{6}{l}{\textbf{Positive support}} \\
OL / Average & 101 & \pione{+1.38}{-2.18,+5.14} & \pione{+2.77}{-4.52,+10.60} & \pione{+2.08}{-1.40,+5.87} & \pione{+4.16}{-3.27,+12.07} \\
OL / TA & 122 & \pione{-0.88}{-6.01,+4.52} & \pione{-4.59}{-14.68,+7.04} & \pione{+0.69}{-4.29,+6.05} & \pione{-1.46}{-11.29,+10.24} \\
QW / Average & 117 & \pione{-0.56}{-2.21,+0.95} & \pione{-1.44}{-3.31,-0.08} & \pione{-0.56}{-2.19,+0.94} & \pione{-1.44}{-3.26,-0.06} \\
QW / TA & 131 & \pione{+0.49}{-3.51,+4.58} & \pione{+1.44}{-3.95,+7.34} & \pione{+0.49}{-3.48,+4.71} & \pione{+1.44}{-3.85,+7.45} \\
Pooled & 471 & \pione{+0.11}{-1.77,+2.05} & \pione{-0.45}{-3.86,+3.37} & \pione{+0.67}{-1.24,+2.62} & \pione{+0.67}{-2.82,+4.46} \\
\addlinespace[3pt]
\rowcolor{RTHeader}
\multicolumn{6}{l}{\textbf{Non-positive support}} \\
OL / Average & 155 & \pione{+0.26}{-4.72,+5.41} & \pione{-2.73}{-11.37,+6.03} & \pione{+0.54}{-4.47,+5.75} & \pione{-2.16}{-11.06,+6.63} \\
OL / TA & 134 & \pione{-0.15}{-4.31,+4.03} & \pione{+5.33}{-1.36,+12.94} & \pione{+2.00}{-1.99,+6.11} & \pione{+9.63}{+3.17,+17.15} \\
QW / Average & 139 & \pione{-2.30}{-4.34,-0.60} & \pione{-1.75}{-5.05,+1.11} & \pione{-1.92}{-4.08,-0.10} & \pione{-0.99}{-4.32,+1.85} \\
QW / TA & 125 & \pione{-0.19}{-4.30,+3.84} & \pione{+0.01}{-4.53,+4.47} & \pione{+0.09}{-2.40,+2.58} & \pione{+0.58}{-3.99,+5.08} \\
\addlinespace[3pt]
\rowcolor{RTHeader}
\multicolumn{6}{l}{\textbf{All events}} \\
OL / Average & 256 & \pione{+0.70}{-2.66,+4.14} & \pione{-0.56}{-6.68,+5.65} & \pione{+1.15}{-2.39,+4.72} & \pione{+0.33}{-5.71,+6.46} \\
OL / TA & 256 & \pione{-0.50}{-3.69,+2.87} & \pione{+0.60}{-5.64,+7.35} & \pione{+1.37}{-1.75,+4.73} & \pione{+4.34}{-1.74,+11.10} \\
QW / Average & 256 & \pione{-1.50}{-2.85,-0.33} & \pione{-1.61}{-3.57,+0.09} & \pione{-1.29}{-2.68,-0.07} & \pione{-1.19}{-3.15,+0.56} \\
QW / TA & 256 & \pione{+0.16}{-2.65,+3.06} & \pione{+0.74}{-2.89,+4.43} & \pione{+0.29}{-2.10,+2.74} & \pione{+1.02}{-2.50,+4.75} \\
\bottomrule
\end{tabular}
\arrayrulecolor{black}
\endgroup
\vspace{-3em}
\end{table*}

\begin{figure}[]
\centering
\includegraphics[width=\textwidth]{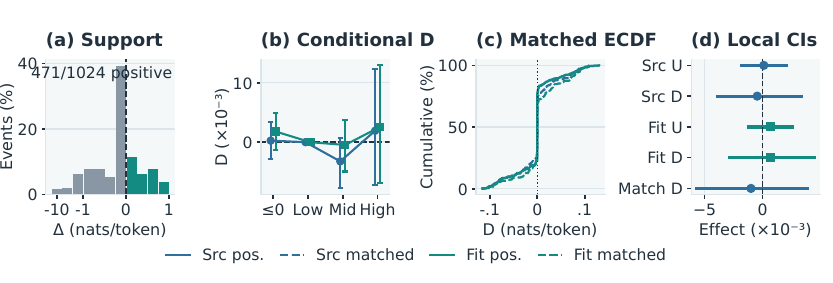}
\caption{\textbf{Local source-supervision diagnostics.} (a) Likelihood advantage across 1,024 events (asinh-spaced axis; labels in nats/token). (b) Directional contrast $D$ by support group. (c) Positive-support and matched-control ECDFs. (d) Local and matched effects with prompt-bootstrap intervals. These are fixed-continuation NLL measurements.}
\label{fig:p1-four-panel-appendix}
\end{figure}

\subsection{Matched support and direction preservation}
\label{app:local-matched-support}

Positive events are matched with replacement to non-positive events from the
same architecture, merge, role, layer, and expert pair, choosing the nearest
native NLL. The 440 matched pairs yield source-direction positive-minus-matched
$U=-0.000341$ nats/token (95\% interval $[-0.003323,0.002631]$) and
$D=-0.000997$ ($[-0.005653,0.003868]$). These contrasts differ from
positive-support means and do not establish enrichment.

A fixed-effect regression of continuous sigmoid weight beyond positive
support has coefficient $+0.0720$ (CI $[-0.0279,0.1985]$). Source and
fitted directions agree in sign on $83.4\%$ of supported events
(CI $[80.0,86.6]\%$), but agreement does not establish local benefit.
Only two of four settings passed the prespecified source-direction sign
screen; the contingent original scale-up was therefore not run.
The empirical distributions and support-bin analysis in
Figure~\ref{fig:p1-four-panel-appendix} describe these fixed local events,
not additional checkpoint trials.

\section{Executed SRR Corrections}
\label{app:srr-execution}

\subsection{Aligned trajectories and numerical checks}
\label{app:execution-protocol}

The execution study uses eight OLMoE/Qwen3-MoE Parent/SRR pairs from the
task evaluation. Each setting has 128 prompts (32 per domain) and one
hash-selected token--pair event per prompt at each of five updated layers,
totaling 5,120 events. Both models are teacher-forced on the same tokens
but produce their own hidden trajectories. The direct and input-response
terms follow Eq.~\ref{eq:srr-main-execution}; they are reconstructed in
FP64 from the specified weights and states.

Inputs agree at the first updated layer. The maximum normalized algebra
residual is $6.9\times10^{-18}$, whereas the maximum native-forward versus
FP64 logit discrepancy is $0.0621$. The latter is a separate measurement,
not an input-response term. Algebraic agreement establishes the stated
accounting identity, not bitwise native replay or the cause of task effects.

\subsection{Complete layer-wise measurements}
\label{app:execution-results}

Tables~\ref{tab:r-olmoe-rms}--\ref{tab:r-qwen3_moe-numeric} retain every
setting and updated layer. Each RMS uses 128 prompts and 10,000 prompt
resamples for pointwise 95\% intervals. Direct, response, and executed RMS
values do not add because their signed components have a cross term.
The response/direct RMS ratios in Figure~\ref{fig:srr-local-diagnostics}
are ratios of reported point estimates, not additive attribution shares.

Native expert-set and dispatch-weight changes are reported separately from
reference-precision logits. A nonzero correction need not change the selected
set. These traces differ from the independent local-NLL sample and the task
items; the cross-setting association with task effects is descriptive, not
an event-wise or causal attribution.

\begin{figure*}[t]
\centering
\includegraphics[width=\textwidth]{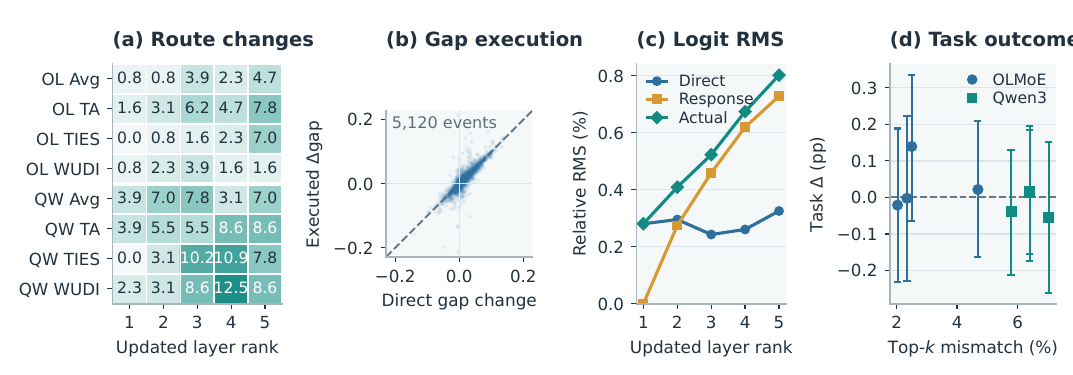}
\caption{\textbf{Fitted and executed corrections.} (a) Native expert-set changes. (b) Direct versus executed pair gaps on 5,120 events. (c) Centered-logit RMS, normalized by each parent before equal-setting averaging. (d) Identity-matched task differences. RMS components are not additive; the task association is descriptive.}
\label{fig:srr-execution-four-panel}
\end{figure*}

\begin{table*}[t]
\centering
\scriptsize
\setlength{\tabcolsep}{4pt}
\renewcommand{\arraystretch}{1.05}
\caption{\textbf{OLMoE SRR pair-gap decomposition.} Each cell is a pair-logit-gap RMS with a pointwise 95\% prompt-bootstrap interval, conditional on its fixed Parent/SRR checkpoints. RMS terms do not add.}
\label{tab:r-olmoe-rms}
\begin{tabular}{@{}llrrr@{}}
\toprule
Merge & Layer & Direct [95\% CI] & Input response [95\% CI] & Executed [95\% CI] \\
\midrule
Avg & 11 & $0.0135\;[0.0116,0.0153]$ & $0\;[0,0]$ & $0.0135\;[0.0116,0.0153]$ \\
 & 12 & $0.013\;[0.0113,0.0147]$ & $0.00341\;[0.00272,0.00415]$ & $0.0135\;[0.0118,0.015]$ \\
 & 13 & $0.0138\;[0.0121,0.0155]$ & $0.00587\;[0.00463,0.00744]$ & $0.0148\;[0.0132,0.0164]$ \\
 & 14 & $0.0151\;[0.0134,0.0168]$ & $0.00835\;[0.00613,0.0109]$ & $0.0189\;[0.0159,0.0222]$ \\
 & 15 & $0.0196\;[0.0165,0.0233]$ & $0.0077\;[0.00648,0.00896]$ & $0.0221\;[0.0187,0.0258]$ \\
\addlinespace[2pt]
TA & 11 & $0.0151\;[0.0127,0.0176]$ & $0\;[0,0]$ & $0.0151\;[0.0127,0.0176]$ \\
 & 12 & $0.0169\;[0.0144,0.0193]$ & $0.0114\;[0.00306,0.0193]$ & $0.0208\;[0.0156,0.027]$ \\
 & 13 & $0.016\;[0.0135,0.0184]$ & $0.0147\;[0.00621,0.0226]$ & $0.023\;[0.0165,0.0304]$ \\
 & 14 & $0.0147\;[0.0131,0.0163]$ & $0.0147\;[0.00697,0.0227]$ & $0.0191\;[0.0156,0.0234]$ \\
 & 15 & $0.0206\;[0.0174,0.0244]$ & $0.0118\;[0.00811,0.0158]$ & $0.0234\;[0.0194,0.0279]$ \\
\addlinespace[2pt]
TIES & 11 & $0.0145\;[0.0123,0.0167]$ & $0\;[0,0]$ & $0.0145\;[0.0123,0.0167]$ \\
 & 12 & $0.0128\;[0.0109,0.0149]$ & $0.00264\;[0.00218,0.00311]$ & $0.0133\;[0.0113,0.0155]$ \\
 & 13 & $0.0146\;[0.0122,0.017]$ & $0.00675\;[0.00452,0.00933]$ & $0.0158\;[0.0134,0.0183]$ \\
 & 14 & $0.0143\;[0.0122,0.0164]$ & $0.00779\;[0.00628,0.00937]$ & $0.0174\;[0.0144,0.0206]$ \\
 & 15 & $0.0228\;[0.0184,0.027]$ & $0.00877\;[0.0069,0.0111]$ & $0.0244\;[0.0198,0.029]$ \\
\addlinespace[2pt]
WUDI & 11 & $0.00697\;[0.00622,0.00771]$ & $0\;[0,0]$ & $0.00697\;[0.00622,0.00771]$ \\
 & 12 & $0.00995\;[0.00825,0.0117]$ & $0.00244\;[0.00193,0.00298]$ & $0.0102\;[0.00856,0.012]$ \\
 & 13 & $0.00693\;[0.00603,0.00783]$ & $0.00713\;[0.00372,0.011]$ & $0.00963\;[0.00773,0.012]$ \\
 & 14 & $0.00903\;[0.00785,0.0102]$ & $0.00886\;[0.00661,0.0113]$ & $0.0131\;[0.0107,0.0157]$ \\
 & 15 & $0.0154\;[0.0136,0.0171]$ & $0.0133\;[0.00875,0.0181]$ & $0.0221\;[0.0175,0.028]$ \\
\bottomrule
\end{tabular}
\end{table*}

\begin{table*}[t]
\centering
\scriptsize
\setlength{\tabcolsep}{4pt}
\renewcommand{\arraystretch}{1.05}
\caption{\textbf{Qwen3-MoE SRR pair-gap decomposition.} Each cell is a pair-logit-gap RMS with a pointwise 95\% prompt-bootstrap interval, conditional on its fixed Parent/SRR checkpoints. RMS terms do not add.}
\label{tab:r-qwen3_moe-rms}
\begin{tabular}{@{}llrrr@{}}
\toprule
Merge & Layer & Direct [95\% CI] & Input response [95\% CI] & Executed [95\% CI] \\
\midrule
Avg & 43 & $0.0323\;[0.0279,0.0365]$ & $0\;[0,0]$ & $0.0323\;[0.0279,0.0365]$ \\
 & 44 & $0.0357\;[0.0303,0.0408]$ & $0.00439\;[0.00319,0.00569]$ & $0.0358\;[0.0304,0.0408]$ \\
 & 45 & $0.0272\;[0.0213,0.0334]$ & $0.00511\;[0.00415,0.00604]$ & $0.0272\;[0.021,0.0339]$ \\
 & 46 & $0.0266\;[0.0233,0.0298]$ & $0.0233\;[0.00859,0.037]$ & $0.035\;[0.0265,0.0452]$ \\
 & 47 & $0.0332\;[0.0275,0.0388]$ & $0.0204\;[0.0134,0.0268]$ & $0.0429\;[0.0339,0.0517]$ \\
\addlinespace[2pt]
TA & 43 & $0.0353\;[0.0303,0.0402]$ & $0\;[0,0]$ & $0.0353\;[0.0303,0.0402]$ \\
 & 44 & $0.0407\;[0.0349,0.0465]$ & $0.0051\;[0.00366,0.00655]$ & $0.0412\;[0.0351,0.0474]$ \\
 & 45 & $0.0266\;[0.0217,0.0313]$ & $0.00824\;[0.00626,0.0101]$ & $0.0274\;[0.0221,0.0323]$ \\
 & 46 & $0.0305\;[0.0267,0.0343]$ & $0.0151\;[0.0109,0.019]$ & $0.0362\;[0.0314,0.0408]$ \\
 & 47 & $0.0493\;[0.0419,0.0566]$ & $0.0185\;[0.0125,0.0254]$ & $0.053\;[0.0456,0.0603]$ \\
\addlinespace[2pt]
TIES & 43 & $0.0227\;[0.0202,0.0252]$ & $0\;[0,0]$ & $0.0227\;[0.0202,0.0252]$ \\
 & 44 & $0.0228\;[0.0201,0.0255]$ & $0.00451\;[0.00302,0.00619]$ & $0.0237\;[0.0207,0.0267]$ \\
 & 45 & $0.0197\;[0.0168,0.0226]$ & $0.0066\;[0.00472,0.00855]$ & $0.0202\;[0.0176,0.0228]$ \\
 & 46 & $0.0218\;[0.0187,0.0249]$ & $0.0107\;[0.0081,0.0134]$ & $0.0231\;[0.0198,0.0264]$ \\
 & 47 & $0.0317\;[0.0282,0.035]$ & $0.0168\;[0.0121,0.0213]$ & $0.0358\;[0.0317,0.0398]$ \\
\addlinespace[2pt]
WUDI & 43 & $0.0308\;[0.0263,0.0351]$ & $0\;[0,0]$ & $0.0308\;[0.0263,0.0351]$ \\
 & 44 & $0.0314\;[0.0266,0.0359]$ & $0.00459\;[0.0032,0.00599]$ & $0.0319\;[0.027,0.0366]$ \\
 & 45 & $0.0283\;[0.0242,0.0325]$ & $0.00933\;[0.00688,0.0117]$ & $0.0294\;[0.0255,0.0334]$ \\
 & 46 & $0.029\;[0.0246,0.0332]$ & $0.0235\;[0.00905,0.0368]$ & $0.038\;[0.0284,0.0485]$ \\
 & 47 & $0.0271\;[0.0231,0.0313]$ & $0.0215\;[0.0149,0.0295]$ & $0.0376\;[0.0295,0.0467]$ \\
\bottomrule
\end{tabular}
\end{table*}

\begin{table*}[t]
\centering
\scriptsize
\setlength{\tabcolsep}{4pt}
\renewcommand{\arraystretch}{1.05}
\caption{\textbf{OLMoE SRR routing and numerical diagnostics.} One hash-selected event per prompt and updated layer. Cross is the mean centered-logit direct--response inner product. The native--FP64 difference is the largest absolute BF16-forward versus reconstructed-logit discrepancy; algebra residual is normalized. Route changes and dispatch-weight distance use native routing.}
\label{tab:r-olmoe-numeric}
\begin{tabular}{@{}llrrrrrr@{}}
\toprule
Merge & Layer & Cross & Top-$k$ change (\%) & Weight $\ell_1$/2 & $\|\Delta h\|/\|h\|$ & Native--FP64 & Algebra resid. \\
\midrule
Avg & 11 & 0 & 0.781 & 0.00042 & 0 & 0.0185 & 0 \\
 & 12 & -1.8e-07 & 0.781 & 0.000836 & 0.00363 & 0.0211 & 0 \\
 & 13 & -6.54e-08 & 3.91 & 0.00198 & 0.00643 & 0.0247 & 0 \\
 & 14 & 3.36e-07 & 2.34 & 0.00173 & 0.00947 & 0.0208 & 0 \\
 & 15 & -1.13e-07 & 4.69 & 0.00255 & 0.0112 & 0.0291 & 0 \\
\addlinespace[2pt]
TA & 11 & 0 & 1.56 & 0.000725 & 0 & 0.0207 & 0 \\
 & 12 & -1.94e-07 & 3.12 & 0.00172 & 0.0045 & 0.0212 & 0 \\
 & 13 & 2.24e-07 & 6.25 & 0.00305 & 0.00841 & 0.0221 & 0 \\
 & 14 & -1.36e-07 & 4.69 & 0.00248 & 0.0117 & 0.0216 & 0 \\
 & 15 & -4.65e-07 & 7.81 & 0.00344 & 0.0133 & 0.0247 & 0 \\
\addlinespace[2pt]
TIES & 11 & 0 & 0 & 0.000249 & 0 & 0.0156 & 0 \\
 & 12 & -3.8e-08 & 0.781 & 0.000761 & 0.00358 & 0.0157 & 0 \\
 & 13 & -2.68e-07 & 1.56 & 0.00135 & 0.00709 & 0.0182 & 0 \\
 & 14 & 3.47e-07 & 2.34 & 0.00195 & 0.00951 & 0.0221 & 0 \\
 & 15 & -8.7e-07 & 7.03 & 0.00345 & 0.012 & 0.0296 & 0 \\
\addlinespace[2pt]
WUDI & 11 & 0 & 0.781 & 0.00033 & 0 & 0.0201 & 0 \\
 & 12 & 6.6e-08 & 2.34 & 0.00122 & 0.00307 & 0.0231 & 0 \\
 & 13 & -8.84e-08 & 3.91 & 0.0018 & 0.0065 & 0.0215 & 0 \\
 & 14 & -1.27e-07 & 1.56 & 0.00141 & 0.0091 & 0.0235 & 0 \\
 & 15 & 2.39e-07 & 1.56 & 0.00172 & 0.011 & 0.0255 & 0 \\
\bottomrule
\end{tabular}
\end{table*}

\begin{table*}[t]
\centering
\scriptsize
\setlength{\tabcolsep}{4pt}
\renewcommand{\arraystretch}{1.05}
\caption{\textbf{Qwen3-MoE SRR routing and numerical diagnostics.} One hash-selected event per prompt and updated layer. Cross is the mean centered-logit direct--response inner product. The native--FP64 difference is the largest absolute BF16-forward versus reconstructed-logit discrepancy; algebra residual is normalized. Route changes and dispatch-weight distance use native routing.}
\label{tab:r-qwen3_moe-numeric}
\begin{tabular}{@{}llrrrrrr@{}}
\toprule
Merge & Layer & Cross & Top-$k$ change (\%) & Weight $\ell_1$/2 & $\|\Delta h\|/\|h\|$ & Native--FP64 & Algebra resid. \\
\midrule
Avg & 43 & 0 & 3.91 & 0.00558 & 0 & 0.0451 & 0 \\
 & 44 & -7.73e-08 & 7.03 & 0.00772 & 0.00338 & 0.0526 & 6.87e-18 \\
 & 45 & -4.01e-07 & 7.81 & 0.00846 & 0.00635 & 0.0476 & 0 \\
 & 46 & -2.88e-07 & 3.12 & 0.00635 & 0.00871 & 0.0543 & 0 \\
 & 47 & -2.05e-07 & 7.03 & 0.00794 & 0.00917 & 0.0613 & 0 \\
\addlinespace[2pt]
TA & 43 & 0 & 3.91 & 0.00417 & 0 & 0.0434 & 0 \\
 & 44 & -3.53e-08 & 5.47 & 0.00691 & 0.00399 & 0.0466 & 0 \\
 & 45 & -1.68e-07 & 5.47 & 0.00779 & 0.00802 & 0.049 & 0 \\
 & 46 & 3.22e-07 & 8.59 & 0.00946 & 0.0107 & 0.0527 & 0 \\
 & 47 & -7.97e-07 & 8.59 & 0.0109 & 0.0115 & 0.0611 & 0 \\
\addlinespace[2pt]
TIES & 43 & 0 & 0 & 0.00108 & 0 & 0.0453 & 0 \\
 & 44 & -1.28e-07 & 3.12 & 0.00365 & 0.00291 & 0.0498 & 0 \\
 & 45 & -8.17e-08 & 10.2 & 0.00958 & 0.00636 & 0.0452 & 0 \\
 & 46 & -3.19e-07 & 10.9 & 0.0101 & 0.00873 & 0.062 & 0 \\
 & 47 & -3.01e-07 & 7.81 & 0.00872 & 0.00965 & 0.0614 & 0 \\
\addlinespace[2pt]
WUDI & 43 & 0 & 2.34 & 0.00333 & 0 & 0.0478 & 0 \\
 & 44 & 8.69e-08 & 3.12 & 0.00437 & 0.00339 & 0.0494 & 0 \\
 & 45 & -1.93e-07 & 8.59 & 0.00901 & 0.00671 & 0.0534 & 0 \\
 & 46 & 3.78e-08 & 12.5 & 0.0138 & 0.0105 & 0.045 & 0 \\
 & 47 & 1.3e-06 & 8.59 & 0.0102 & 0.0113 & 0.0618 & 0 \\
\bottomrule
\end{tabular}
\end{table*}

\providecolor{EvalFHeader}{HTML}{E9F1F2}
\providecolor{EvalFRow}{HTML}{F6F8F9}

\section{Supplementary Task Evaluation}
\label{app:paired-evaluation}

\paragraph{Scoring and uncertainty.}
\label{app:task-evaluation-protocol}
Each Parent/SRR pair is evaluated five times on 35,326 matched items using
five-shot \texttt{lm-evaluation-harness} with a VLLM backend.
Table~\ref{tab:benchmark-scoring} lists the task sizes and scoring fields.
The aggregate equally weights the eight task percentages.
For each matched item, we first average the paired SRR--Parent score
difference over the five evaluation runs. Paired item bootstraps then
resample item identities within each task and recompute both task
differences and the aggregate.
Intervals condition on fixed checkpoints; they do not capture variation
across repair seeds, calibration samples, or training runs.
The SRR--Parent intervals do not apply to HARC--Parent or SRR--HARC comparisons.

\begin{table}[!htbp]
\centering
\small
\caption{Evaluation items and scoring fields. Task scores receive equal weight.}
\label{tab:benchmark-scoring}
\begingroup
\setlength{\tabcolsep}{5pt}
\renewcommand{\arraystretch}{1.10}
\begin{tabularx}{\textwidth}{@{\hspace{5pt}}l r >{\raggedright\arraybackslash}X@{\hspace{5pt}}}
\toprule
\rowcolor{EvalFHeader}
\textbf{Benchmark} & \textbf{Items} & \textbf{Scoring field} \\
\midrule
MMLU & 14,042 & \texttt{acc,none} \\
HellaSwag & 10,042 & \texttt{acc\_norm,none} \\
ARC-Challenge & 1,172 & \texttt{acc\_norm,none} \\
ARC-Easy & 2,376 & \texttt{acc\_norm,none} \\
PIQA & 1,838 & \texttt{acc\_norm,none} \\
WinoGrande & 1,267 & \texttt{acc,none} \\
BoolQ & 3,270 & \texttt{acc,none} \\
GSM8K & 1,319 & \texttt{exact\_match,flexible-extract} \\
\bottomrule
\end{tabularx}
\endgroup
\end{table}

\begin{samepage}
\paragraph{Supplementary selection test.}
\label{app:independent-candidates}
A separate pilot tested 40 nonzero source, opposite, and random-direction
proposals across four settings. Multiplicity-controlled choice-loss tests
and task guardrails preceded evaluation on 1,536 confirmation items.
All 20 family--setting decisions retained Parent, yielding zero confirmation
differences by checkpoint identity. Fixed, nonselected source-scale-one
updates yielded $-0.016$ pp (95\% paired CI $[-0.179,+0.146]$).
Rejection under this rule does not establish that every candidate is ineffective.
\par
\end{samepage}

\end{document}